\documentclass{article}

\usepackage[main, preprint]{neurips_2026}
\usepackage{graphicx} 

\usepackage[utf8]{inputenc}
\usepackage[T1]{fontenc}
\usepackage{hyperref}
\usepackage{url}
\usepackage{booktabs}
\usepackage{array} 
\usepackage{amsmath}
\usepackage{amsfonts}
\usepackage{amssymb}
\usepackage{nicefrac}
\usepackage{microtype}
\usepackage{xcolor}
\usepackage{algorithm}
\usepackage{algpseudocode}
\usepackage{enumitem}
\usepackage{placeins} 
\newlist{todolist}{itemize}{2}
\setlist[todolist]{label=$\square$}

\title{An Educated Guess: Deriving Statistically Aligned Gradient Estimates for Zeroth-Order Optimisation}

\author{%
  Justus Hübotter \\
\And
  Serge Thill \\
\And
  Marcel van Gerven \\
\And
  Nasir Ahmad \\
  Donders Institute for Brain, Cognition, and Behaviour\\Radboud University, Nijmegen, the Netherlands\\
  \texttt{\{justus.huebotter, nasir.ahmad\}@ru.nl} \\
}

\begin{document}

\maketitle

\begin{abstract}
As novel hardware architectures are explored for the purposes of machine learning, including neuromorphic chips, asynchronous systems, and ASICs, effective learning algorithms are highly desirable.
While backpropagation is the standard for differentiation on standard hardware, its requirement for global synchrony and exact differentiability limits the exploration of exotic algorithms on non-traditional substrates and devices.
Zeroth-order (ZO) optimisation methods, such as SPSA and weight perturbation, offer a compelling alternative by estimating gradients through inference-only passes; however, these methods have historically suffered due to the `curse of dimensionality', where random perturbations become increasingly misaligned with the true gradient in high-dimensional spaces.
In this work, we move beyond blind random noise by deriving zeroth-order search directions from activation geometry available during the forward pass.
This leads to a family of Spectral Zeroth-Order methods, SZO-\(\kappa\), whose directions solve local response-based gradient-guessing objectives under explicit stochastic assumptions.
These methods explore feature-supported directions that are more likely to produce informative loss responses.
Using multilayer and convolutional networks, we analyse gradient alignment, bias-variance structure, and training stability of these methods relative to standard perturbation baselines and backpropagation.
\end{abstract}

\section{Introduction}
Backpropagation is the workhorse of modern deep learning~\citep{rumelhart_learning_1986,werbos1990backpropagation,baydin2018automatic}.
Its success, however, comes with a specific computational constraint: the model must be described through a differentiable graph, with stored or reconstructable intermediate state, and a backward computation that is structurally different from inference.
This is acceptable on conventional multipurpose hardware, but it is less natural for neuromorphic and physical substrates, asynchronous or event-driven systems, networks with non-differentiable components, and memory-limited settings~\citep{roy2019towards,schuman2022opportunities,gu_efficient_2021,VanGerven2025Neuromorphic}.
Forward-only and local-credit alternatives have therefore been explored as ways to avoid global backpropagation~\citep{silver_learning_2021,baydin_gradients_2022,hinton2022forward}.
Recent interest in forward-only training for large models gives the same issue a second, more practical motivation: if learning can be reduced to inference-like evaluations, the memory footprint and hardware applicability improves substantially~\citep{malladi_fine-tuning_2023,zhang_revisiting_2024}.

Zeroth-order (ZO) optimisation and perturbation learning propose a solution.
Instead of differentiating a model, they perturb parameters or activities, observe the change in a scalar objective, and correlate that change with the injected perturbation.
This idea appears in stochastic approximation, finite-difference ZO methods, simultaneous perturbation stochastic approximation (SPSA), weight perturbation, node perturbation, and evolution strategies~\citep{kiefer_stochastic_1952,barto_neuronlike_1983,mazzoni_more_1991,jabri_weight_1991,spall_multivariate_1992,cauwenberghs_fast_1992,seung_learning_2003,werfel_learning_2003,fiete_gradient_2006,legenstein_reward-modulated_2010,salimans_evolution_2017,such_deep_2018,nesterov_random_2017}.
The attraction of being able to use the same computational steps as inference but instead for training is significant.
Not to mention the applicability of these methods to systems in which the exact derivatives are expensive to compute or even unknown.
There also, however, exists an equally clear obstacle: randomly searching through the parameter space of a model scales terribly in high dimensions.
Any perturbations or search directions are usually a bad gradient guess in high dimensions and therefore the estimated gradients have high variance (noise). 
These sources of error are a central reason that classical perturbation learning becomes unstable or sample-inefficient as networks widen, deepen, or share parameters~\citep{flaxman_online_2004,ghadimi_stochastic_2013,shamir_optimal_2015,nesterov_random_2017,hiratani_stability_2022,zuge_weight_2023}.

This paper asks whether perturbation learning can choose better search directions using information already present during the forward pass.
Our starting point is that neural network layers have local feature geometry: inputs to a layer are not uniformly distributed in all directions, and perturbations that respect this structure should produce more informative loss responses than blind isotropic noise.
We formalise this observation as a local stochastic gradient-guessing problem and \textbf{we derive a family of Spectral Zeroth-Order estimators, SZO-$\kappa$, which are explicitly derived as a solution to the ideal gradient search direction if one has only local information from the forward-pass}.

This SZO-$\kappa$ family also captures standard weight perturbation on one end and recovers node-like perturbations at the other end of a particular choice in $\kappa$.
This perspective also gives a practical bridge to new node-level perturbation methods, where activity-level credit is induced by realisable parameter noise rather than by independently perturbing every node.
After deriving SZO-\(\kappa\), we collect some practical stabilisers from the literature that address its remaining finite-sampling errors (Appendix~\ref{app:corrections}).
This connects these corrections to established ideas in gradient guessing, preconditioning, and natural-gradient learning~\citep{amari_natural_1998,martens_optimizing_2015,ren_scaling_2022, singhal_how_2023,vyas_soap_2025}, while keeping the main contribution focused on feature-geometric zeroth-order search.

\section{Zeroth-Order Gradient Estimators}

\subsection{Setup and Notation}
Let \(f(x,\theta)\) denote an arbitrary parametric function with flattened parameters \(\theta\in\mathbb{R}^m\), prediction \(\hat{y}=f(x,\theta)\), and scalar loss \(\mathcal{L}(\hat{y},y)\). For compactness we write \(\mathcal{L}(\theta)\) when \(x\) and \(y\) are fixed, and \(g=\nabla_\theta\mathcal{L}(\theta)\) for the row-gradient. Backpropagation computes \(g\) by differentiating the computational graph using reverse accumulation; zeroth-order methods instead estimate it from changes in the scalar loss after perturbing the system, which may be a neural network, a recurrent or physical system, or a module containing non-differentiable components.
Estimating gradients well is therefore heavily dependent on the statistical design of search directions~\citep{kiefer_stochastic_1952,flaxman_online_2004,ghadimi_stochastic_2013,shamir_optimal_2015,nesterov_random_2017}.

A probe refers to one perturbed forward evaluation along with the (one) corresponding scalar loss observation. We focus throughout on one-sided estimators (a clean baseline plus \(n\) perturbed evaluations), since they double the number of directions explored per compute budget relative to two-sided SPSA-style schemes. Perturbations may be applied simultaneously across layers or layer-by-layer~\citep{dalm_effective_2023}, and credit may be assigned globally or per-example when sample-wise losses are available.

\subsection{Finite-Difference Approaches}
For a perturbation direction \(\psi\in\mathbb{R}^m\) and scale \(\epsilon>0\), a small-step Taylor expansion gives
\begin{equation}\label{eq:FO}
     \mathcal{L}(\theta+\epsilon\psi)-\mathcal{L}(\theta)
  = \epsilon g\psi + O(\epsilon^2), 
\end{equation}
so a single forward difference measures the directional derivative, \(g\psi\). Given sampled directions \(\Psi_{1:n}=\{\psi_i\}_{i=1}^n\), the one-sided estimator and its empirical covariance are
\[
  \widehat{g}_n
  =
  \frac{1}{n\sigma^2\epsilon}\sum_{i=1}^n
  \big(\mathcal{L}(\theta+\epsilon\psi_i)-\mathcal{L}(\theta)\big)\psi_i^\top,
  \qquad
  M_n=\frac{1}{n\sigma^2}\sum_{i=1}^n\psi_i\psi_i^\top .
\]
Here \(\sigma^2\) is the per-coordinate perturbation variance; in all experiments we take \(\sigma=1\), so that \(\epsilon\) is the per-coordinate perturbation standard deviation. Substituting the linearisation, Equation~\ref{eq:FO}'s first term, gives \(\widehat{g}_n\approx gM_n\), and for iid isotropic directions \(\mathbb{E}_{\psi}[M_n]=I_m\), so the population estimator recovers \(g\) up to the finite-difference remainder.
For any finite \(n\) the error is controlled by how close \(M_n\) is to \(I_m\). When the sampled directions, \(\psi_i\), are iid isotropic over all \(m\) parameters, this estimator is weight perturbation (WP)~\citep{jabri_weight_1991,cauwenberghs_fast_1992,werfel_learning_2003}; the two-sided variant is SPSA~\citep{spall_multivariate_1992}. Both treat every parameter direction as equally plausible.

\subsection{Bias, Variance, and the Curse of Dimensionality}
For a finite number of probes \(n\), there is a deviation of $M_n$ from identity: \(M_n = I_m + E_n\), where \(E_n\) is some error. The linearised estimator therefore satisfies \(\widehat{g}_n \approx g + gE_n\). 
The total mean squared estimation error then decomposes into two terms such that,
\[
  \mathbb{E}_{\psi}\!\left[\|\widehat{g}_n - g\|_2^2\right]
  =
  \mathbb{E}\!\left[\|g E_n\|_2^2\right] + \mathcal{O}(\epsilon^2).
\]
For probes with iid zero-mean, unit-variance coordinates (e.g.\ Gaussian, uniform, or Rademacher perturbations), \(E_n=\frac{1}{n}\sum_i(\psi_i\psi_i^\top-I_m)\) is an average of \(n\) independent mean-zero terms, and a direct fourth-moment calculation gives
\[
  \mathbb{E}\big[\|gE_n\|_2^2\big] = \frac{m+\mu_4-2}{n}\,\|g\|_2^2 ,
\]
where \(m\) is the number of parameters, \(n\) the number of probes, and \(\mu_4=\mathbb{E}[\psi_{ij}^4]\) the fourth moment of a probe coordinate (\(\mu_4=3\) for Gaussian, \(9/5\) for uniform, and \(1\) for Rademacher probes). Consequently,
\[
  \mathbb{E}_{\psi}\!\left[\|\widehat{g}_n-g\|_2^2\right]
  =
  \mathcal{O}\!\left(\frac{m}{n}\right)\|g\|_2^2 + \mathcal{O}(\epsilon^2).
\]
This shows the `curse of dimensionality' for perturbation learning. Specifically, maintaining a given estimation error requires the number \(n\) to grow with \(m\). In wide networks much of this search is spent on directions with little effect on the current input, producing a blind diffusion of parameters with WP~\citep{zuge_weight_2023}.

The classical response is to perturb at a different locus.
Node perturbation (NP) injects noise into a layer's activity \(h\in\mathbb{R}^k\) rather than its parameters, estimating loss sensitivity at the activations and assigning credit through the local Jacobian. For a linear pre-activation \(h = Wa\) with input activity \(a\in\mathbb{R}^q\), an injected perturbation \(\phi\in\mathbb{R}^k\) yields a perturbed output \(\tilde{h} = Wa + \phi\).
Similarly to weight perturbation, directional exploration across sampled perturbations \(\{\phi_i\}_{i=1}^n\) estimates the activation gradient \(\nabla_h\mathcal{L}\) via
\[
  \widehat{\nabla_h\mathcal{L}}_n
  =
  \frac{1}{n\sigma_\phi^2\epsilon}\sum_{i=1}^n
  \big(\mathcal{L}(h + \epsilon\phi_i) - \mathcal{L}(h)\big)\phi_i^\top,
\]
which satisfies \(\mathbb{E}[\widehat{\nabla_h\mathcal{L}}_n] = \nabla_h\mathcal{L} + \mathcal{O}(\epsilon)\) for isotropic \(\phi_i\). By the chain rule, the gradient with respect to the weights is \(\nabla_W \mathcal{L} = (\nabla_h \mathcal{L})^\top a^\top\). Substituting the activation estimate yields the weight update:
\[
  \widehat{\nabla_W\mathcal{L}}_n
  =
  \widehat{\nabla_h\mathcal{L}}_n^\top a^\top
  \approx
  \frac{1}{n\sigma_\phi^2\epsilon}\sum_{i=1}^n \Delta\mathcal{L}_i \, \phi_i a^\top,
\]
where \(\phi_i a^\top\) forms a rank-one outer product corresponding to the transpose Jacobian \(J_{h,\mathrm{vec}(W)}^\top\phi_i\)~\citep{barto_neuronlike_1983,mazzoni_more_1991,seung_learning_2003,werfel_learning_2003,fiete_gradient_2006}. 

Because exploration is confined to the activation manifold \(\mathbb{R}^k\) rather than the parameter space \(\mathbb{R}^{k \times q}\), the estimation variance depends on the activation dimension \(k\) instead of the parameter count \(m = kq\). For sample-wise losses, the resulting error scales as
\[
  \mathbb{E}\!\left[\|\widehat{\nabla_h\mathcal{L}}_n - \nabla_h\mathcal{L}\|_2^2\right]
  =
  \mathcal{O}\!\left(\frac{k}{n}\right)\|\nabla_h\mathcal{L}\|_2^2 + \mathcal{O}(\epsilon^2),
\]
which translates to an estimation variance reduction of factor \(q\) over weight perturbation -- a substantial improvement in wide, overparameterised layers.
However, this dimensional advantage weakens under aggregate batch rewards, simultaneous multi-layer perturbation~\citep{hiratani_stability_2022}, or extensive weight sharing (such as in convolutions), where the effective ratio of parameters to nodes is altered.

This covariance view separates two kinds of error: linearisation error, affected by the finite-difference scale \(\epsilon\), and search-direction bias and variance, controlled by the statistics of \(\Psi\). Since \(M_n\) is positive semidefinite, a biased search does not flip the descent direction, but it can leave the estimate weakly aligned, high variance, or nearly blind to important directions, which motivates shaping \(M_n\).

\section{Spectral Zeroth-Order Perturbations}

\subsection{Gradient Searches Should Align with Feature Geometry}
One way to pose the core question of a zeroth-order method is to ask: how should a search direction be selected which is most likely to return the gradient?
In this work, we consider answering this question for a neural network layer with observed inputs.
This shares the conceptual goal of Covariance Matrix Adaptation~\citep{hansen_completely_2001}, but explicitly constructs the search geometry from forward-pass feature statistics rather than historical search paths.

Consider a single layer with weights \(W\in\mathbb{R}^{k\times q}\). For a mini-batch of \(B\) examples, one may collect the layer inputs during a forward pass as columns of \(A=[a_1,\ldots,a_B]\in\mathbb{R}^{q\times B}\). For convolutional layers, \(A\) collects the flattened receptive-field patches across all spatial positions and batch elements, treating each patch as an independent example. The pre-activations produced by the layer are
\[
  H = WA \in \mathbb{R}^{k\times B},
\]
where column \(h_j=Wa_j\) is the activity for example \(j\) (or patch \(j\)). The remaining downstream network maps \(H\) to the batch loss \(\mathcal{L}_B\). We assume, as is common for ZO methods, that the downstream computation is differentiable at the current \(H\), and perturbations are small enough that the loss change is well approximated by its linearisation.

For analysis, let \(V:=\nabla_H\mathcal{L}_B\in\mathbb{R}^{k\times B}\) denote the downstream error signal at this layer.
We shall not make use of \(\nabla_H\mathcal{L}_B\) directly, but it is informative to introduce it to explore the problem of finding gradient directions.
With this notation, the layer gradient factors as
\[
  \nabla_W\mathcal{L}_B =  \nabla_H\mathcal{L}_B A^\top = VA^\top.
\]
Thus, the gradient is not arbitrary in \(\mathbb{R}^{k\times q}\), in truth its right-hand geometry is determined by the batch activations \(A\). From another perspective, one can observe that a weight perturbation \(\Psi\in\mathbb{R}^{k\times q}\) induces the activity perturbation \(\Delta H=\Psi A\), so directions orthogonal to the row space of \(A\) use perturbation budget without changing the layer output.
Our contribution is to turn this observation into an explicit spectral perturbation family for realisable weight noise, with \(\kappa\) controlling the tradeoff between isotropic coverage and feature-concentrated response.

This observation allows us to turn zeroth-order learning away from a blind random search using information available locally during a forward pass.
In particular, even before the downstream error \(V\) is known, the forward pass reveals which parameter directions can causally move the network's response to the current data.
We therefore ask the question: of all possible weight perturbations with a fixed norm \(\|\Psi\|_F^2=c\), which directions maximise the squared change in the loss?
Formally, this is the problem of gradient descent and can be written as
\[
  \Psi^\star = \arg\max_{\|\Psi\|_F^2=c} \langle \nabla_W\mathcal{L}_B,\Psi\rangle^2_F = \arg\max_{\|\Psi\|_F^2=c} \langle \nabla_H\mathcal{L}_B,\Delta H\rangle^2_F = \arg\max_{\|\Psi\|_F^2=c} \langle V,\Delta H\rangle^2_F
\]
Maximising this squared directional derivative is equivalent to finding the true gradient direction, i.e. the direction of steepest descent.
Since the exact downstream error \(\nabla_H\mathcal{L}_B\) is unknown during the forward pass, we assume that \(\nabla_H\mathcal{L}_B\) is isotropic and revisit this assumption later.
The expected squared loss response would therefore, under the assumption of an isotropic downstream network sensitivity, be directly proportional to the induced activity energy \(\|\Delta H\|_F^2 = \mathrm{Tr}(\Psi A A^\top\Psi^\top) = \mathrm{Tr}(\Psi C\Psi^\top)\).
In other words, maximising the activity change at the output of this layer would be the best guess for finding a gradient direction!

More formally,
\[
  \mathbb{E}_V\!\left[ \Delta\mathcal{L}_B^2 \right] \approx \epsilon^2 \mathbb{E}_V\!\left[\langle V A^\top,\Psi\rangle_F^2\right]
  \propto
  \mathrm{Tr}(\Psi C\Psi^\top),
  \qquad C=AA^\top .
\]
Here \(C\in\mathbb{R}^{q\times q}\) is the uncentred activation second-moment matrix. Appendix~\ref{app:response-surrogates} provides in greater detail the isotropic downstream-gradient assumption that justifies this as our surrogate optimisation.
Under this local surrogate, the response-maximising, and therefore best gradient search-direction problem reduces to a constrained optimisation problem, namely
\[
  \Psi^\star
  =
  \arg\max_{\|\Psi\|_F^2=c}\mathrm{Tr}(\Psi C\Psi^\top).
\]

This problem is optimised when \(\Psi C=\lambda\Psi\), which corresponds to the case in which every row of our perturbation lies in an eigenspace of the matrix \(C\), returning a response \(\mathrm{Tr}(\Psi C\Psi^\top)=\lambda c\).
To specifically maximise this optimisation problem, and to have our layer respond maximally to our perturbation, we must therefore place the whole budget \(c\) on the leading eigenvector \(u_1\) of \(C\) (with largest eigenvalue \(\lambda_1\)), giving the rank-1 perturbation
\[
  \Psi^\star=\sqrt{c}\,r u_1^\top,
  \qquad \|r\|_2=1,
\]
with response \(c\lambda_1\) (Appendix~\ref{app:matched-filter}).
The output-side direction \(r\in\mathbb{R}^k\) is left free to be chosen and randomising it isotropically corresponds to the SZO-\(\infty\) estimator that we show below.

\subsection{From Rank-1 Search to Matched Filters}
The rank-1 solution is a consequence of the squared Frobenius energy \(\mathrm{Tr}(\Psi C\Psi^\top)=\|\Psi A\|_F^2=\sum_i s_i(\Psi A)^2\) rewarding singular values quadratically.
The same induced-activity energy can be evaluated under a reweighting of its singular spectrum: replacing every \(s_i(\Psi A)^2\) by \(s_i(\Psi A)\) gives \(\sum_i s_i(\Psi A) = \mathrm{Tr}((\Psi C\Psi^\top)^{1/2}) = \|\Psi A\|_*\), the nuclear norm.
In this way, every induced singular direction is valued linearly rather than quadratically, encouraging rank coverage instead of only focusing on a single direction.
Under the same budget, \(\|\Psi\|_F^2=c\), the corresponding matched-filter problem is
\[
  \Psi^\star
  =
  \arg\max_{\|\Psi\|_F^2=c}\mathrm{Tr}((\Psi C\Psi^\top)^{1/2}),
\]
which provides a solution such that $\Psi^\star \propto R C^{1/2}$ where \(R\) has orthonormal columns (exact for \(k\ge q\); an isotropic random \(R\) satisfies this in expectation, \(\mathbb{E}[R^\top R]=I\)) and \(C^{1/2}\) shapes each draw by the square root of the activation spectrum (see Appendix~\ref{app:matched-filter} for both the rank-1 and matched-filter derivations).

\subsection{The SZO-\texorpdfstring{\(\kappa\)}{kappa} Family}
Having derived the rank-1 and the matched-filter solutions, we can bring these together under a single family of solutions.
We can do so by defining Spectral Zeroth-Order estimation, SZO-\(\kappa\), which takes a random output-side matrix \(R\in\mathbb{R}^{k\times q}\) and shapes its input-side spectrum by the activation second-moment matrix:
\[
  \Psi_\kappa = \sqrt{\frac{c}{\mathrm{Tr}(C^\kappa)}}\,R C^{\kappa/2}.
\]
The factor at the beginning fixes the expected size of the perturbation (\(\mathbb{E}\|\Psi_\kappa\|_F^2=c\), matching the budget, independent of \(\kappa\) and of the scale of the inputs), the matrix \(R\) is a mean-zero isotropic random matrix with iid entries of variance \(1/k\) (e.g.\ Gaussian), so that \(\mathbb{E}[R^\top R]=I_q\), and $C$ is our previously defined activation second-moment matrix. The parameter 
\(\kappa\) directly controls how much more strongly we search along high-variance feature directions than low-variance ones: along the eigendirection \(u_j\) of \(C\), the perturbation variance is proportional to \(\lambda_j^\kappa\).
The resulting finite-difference estimator is still forward-only, but it is no longer `blind' to the input data; instead it is intentionally aligned with the feature geometry revealed by the current batch.
The power family is a natural way to interpolate between these solutions: rescaling the inputs multiplies every power of \(C\) by a single constant, which the normaliser removes, so the relative emphasis on each feature direction does not depend on the scale of the inputs. With the scale fixed by the \(\mathrm{Tr}(C^\kappa)\) normaliser, \(\kappa\) acts purely as a bias--variance dial: \(\kappa=0\) is unbiased isotropic random search (equivalent to WP), \(\kappa=1\) is a matched filter which equalises signal-to-noise across feature directions, larger \(\kappa\) trades shrinking variance for growing bias (Appendix~\ref{app:bias-variance-tradeoffs}), and \(\kappa\to\infty\) collapses to a rank-1 search along the leading eigenvector \(u_1\) of \(C\).
Because the bias is explicit and built from the known input geometry, it can be handled after the fact: whitening by the input factor recovers an unbiased gradient estimate, or the bias can be kept and treated as a preconditioned step, which optimisers such as Shampoo/SOAP or Muon then account for (Appendices~\ref{app:corrections} and~\ref{app:whitening-corrections}).

In this work, we specifically consider, compare and simulate \(\kappa=0\), \(\kappa=1\) and \(\kappa=\infty\) for the SZO family, all of which have interpretable relationships to existing perturbation methods.
A sweep over \(\kappa\in\{0,0.1,\ldots,1,2,5,10,\infty\}\) (Appendix~\ref{app:kappa-sweep}) shows that alignment with backpropagation improves steadily from \(\kappa=0\) to \(\kappa=1\) and saturates by \(\kappa\approx2\).

\paragraph{Equivalences between SZO and other Perturbation Methods}
The SZO family recovers the standard perturbation methods in two key cases, and a recently proposed novel method in a third:
\begin{itemize}
  \item \textbf{SZO-\(0\) \(\;\Leftrightarrow\;\) WP.} With \(C^0=I\), \(\Psi_0\propto R\) is isotropic parameter noise: a random search without preferred directions. The limit \(\kappa\to0^{+}\) differs only for rank-deficient \(C\) (Appendix~\ref{app:equivalences}).
  \item \textbf{SZO-\(1\) \(\;\Leftrightarrow\;\) Node Through Weight Perturbation (NTWP).} The perturbation rule introduced as a heuristic by~\citet{zuge_weight_2023} pairs WP's isotropic weight perturbation with NP-style credit assignment via the induced activity perturbation. This was originally referred to as a `hybrid' perturbation. Under per-batch credit, the NTWP estimator recovers SZO-\(1\) in expectation (\(\propto GC\)) without requiring a matrix square root. See Appendix~\ref{app:equivalences} for the derivation and full algorithm. 
  \item \textbf{{SZO-\(\kappa\) (with \(\kappa > 0\)) \(\;\Leftrightarrow\;\) sample-wise NP}}
For a single sample \(A = a\), the matrix \(C = aa^\top\) is rank-1 with leading eigenvalue \(\|a\|_2^2\) and unit eigenvector \(u = a/\|a\|_2\). For any \(\kappa > 0\), the matrix power is simply the renormalised rank-1 projector \(C^{\kappa/2} = \|a\|_2^\kappa u u^\top\). Because \(\sqrt{\mathrm{Tr}(C^\kappa)} = \|a\|_2^\kappa\), the scale \(\|a\|_2^\kappa\) cancels identically out of the SZO-\(\kappa\) rule:
\[
  \Psi_\kappa 
  = \sqrt{\frac{c}{\mathrm{Tr}(C^\kappa)}} R C^{\kappa/2} 
  = \sqrt{c} \, R u u^\top 
  = \frac{1}{\|a\|_2} \phi u^\top 
  = \frac{1}{\|a\|_2^2} \phi a^\top,
\]
where \(\phi := \sqrt{c}\,\|a\|_2\,Ru \in \mathbb{R}^k\) is an isotropic node-wise perturbation vector. The induced pre-activation perturbation is therefore \(\Delta h = \Psi_\kappa a = \phi\), and the parameter update scales as \(\Delta\mathcal{L} \, \phi a^\top\), recovering exactly Node Perturbation for all \(\kappa > 0\) when done on a per-sample basis.
\end{itemize}

Note, these equivalences are only exact under specific conditions.
The SZO-\(\kappa\)/NP correspondence holds for linear layers in the sample-wise condition.
Convolutional weight sharing breaks the SZO-\(\kappa\)/NP correspondence as even a single image yields a high-rank \(C\), and a shared kernel perturbation correlates node perturbations across positions.
Finally, the SZO-\(1\) correspondence only holds in expectation (see Appendix ~\ref{app:equivalences}) rather than per update, and only under batch-scalar (whole-batch) loss differences: with per-sample credit, as in our training experiments, NTWP is instead an unbiased, NP-like estimator of the gradient (Appendix~\ref{app:equivalences}).

To avoid the computational cost of computing eigenvectors of \(C\) we use (SZO-\(\infty\)) in a sample-wise mode where we treat each batch element as an individual sample, see Appendix~\ref{iswp_details} for more details on this. For convolutional layers this means SZO-\(\infty\) is implemented via the energy-weighted patch surrogate of Appendix~\ref{iswp_details}, i.e.\ an approximation to the exact leading-eigenvector construction rather than the exact estimator.

\subsection{Batch vs.\ Sample Level Loss Measurement}
\label{sec:credit-granularity}
A probe yields one scalar loss difference for the whole batch (batch-scalar
credit) or one difference per sample (per-sample credit), and the two
regimes differ by two orders of magnitude in feedback: at our default batch
size \(B=128\), per-sample methods observe \(128\) loss differences per probe
where batch-scalar methods observe one. Which regime a method can operate in is a
property of its eligibility and implementation, such that:
\begin{itemize}
  \item \textbf{NP, NTWP, and SZO-\(\kappa\)} all admit per-sample or per-batch credit. NP simply samples noise either once and repeats it across batch elements or samples it independently for each batch element; per-batch credit then uses the batch-mean loss difference. SZO-\(\kappa\) methods require the computation of covariance matrices and eigenvector directions which can in some cases (e.g. \(\kappa=\infty\) or when in per-sample mode for linear layers) reduce to simple solutions.
  \item \textbf{WP} methods are often applied at the batch-level by construction (one
  shared perturbation, one scalar response); per-sample weight-perturbation
  variants exist but are extremely memory expensive and we therefore do not simulate these except where absolutely necessary.
\end{itemize}

Where necessary, we provide indications within figures as to whether per-batch or per-sample applications of these methods are being used.


\section{Experiments and Results}

We begin by analysing the gradient alignment and training accuracy differences of various zeroth-order methods while varying network structure and tasks.
We train a two-layer MLP on Fashion-MNIST~\citep{xiao_fashion-mnist_2017} and a six-layer CNN (five convolutional layers and a linear readout) on CIFAR-10~\citep{krizhevsky2009learning}, both with leaky ReLU activations; the full architectures are given in Appendix~\ref{app:architectures}.
All models are trained with cross-entropy loss and label smoothing of $0.1$.
We compare backpropagation (BP), node perturbation (NP), weight perturbation (equivalent to SZO-0), node through weight perturbation (NTWP), and SZO-$\kappa$ at $\kappa\in\{1, \infty\}$.
Each perturbation method is run in \emph{iterative} mode, in which layers are perturbed one at a time at up to $L\times$ forward-pass cost to remove inter-layer crosstalk and produce a higher-fidelity gradient estimate~\citep{dalm_effective_2023} (see Appendix~\ref{app:corrections}).
All configurations are optimised with SGD for $250$ epochs; the learning rate of each method is selected on a held-out validation split, and the selected configuration is then retrained with new seeds to report final test performance (see Appendix~\ref{app:compute_hyperparams} for compute resources and full hyperparameter settings), without activation awareness (Appendix~\ref{app:activation-awareness}).
Code is available at \url{https://github.com/EfficientNeuralSystems/spectral-zeroth-order}.

\begin{figure}
    \centering
    \includegraphics[width=\linewidth]{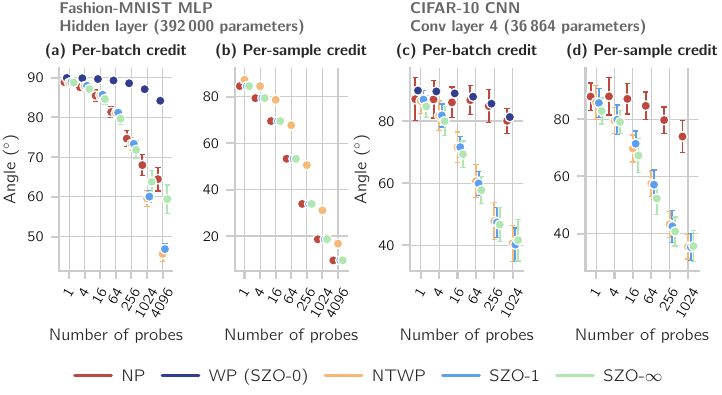}
    \caption{\textbf{The angles between estimated gradients computed by NP, WP, SZO-\(\kappa\) and NTWP against the true backpropagation gradient for randomly initialised layers in MLP and Convolutional Networks.}
    For each network, the left panel uses per-batch loss-difference credit and the right panel per-sample credit; points show the mean angle over 20 randomly seeded initialisations (batch of \(128\)) against the number of probes \(n\), with error bars of \(\pm 1\) standard deviation.
    Left pair, the gradient alignment angle is shown for the first layer of our Fashion-MNIST MLP network.
    Under per-sample credit these angles clearly show the equivalence of NP and SZO-\(\kappa\) on a linear layer, as theorised in this work.
    Right pair, the gradient alignment angle is shown for the last \(3\times3\) convolutional layer (\(64\to64\) channels) of our CIFAR-10 CNN model.
    On convolutional layers NP suffers significantly from its lack of shared perturbation across kernel positions.
    Note, these angles were measured by perturbing only the layer in question and leaving all other network layers unperturbed.
    }
    \label{fig:angle_alignments}
\end{figure}

Figure~\ref{fig:angle_alignments} demonstrates how well gradient estimates formed by NP, WP, NTWP, and the SZO-family align with that computed by backpropagation.
This is shown for a linear and convolutional layer on two different tasks.
Most notably, one can see that despite the number of passes being orders of magnitude lower than the number of parameters (or nodes) in these layers, SZO-based gradient estimators (as well as NTWP and sometimes NP) often reach levels of alignment which far exceed expectations.
We attribute this effect to the manner in which these methods can align their updates to the geometry of the feature data at these layers, and that the effective dimensionality to search over is then far lower than the dimension of the ambient parameter space.

A number of confirmations of our theoretical derivations above are also demonstrated in these alignment measurements.
Namely, as predicted by our theoretical derivations: SZO-1 and NTWP are indistinguishable in their alignment for ``per-batch'' credit assignment, NP is indistinguishable from all SZO methods in linear layers when using per-sample credit assignment, and SZO (as well as NTWP) methods beat NP in their alignment to BP when applied to convolutional layers.
Some interesting features to note: in per-batch credit assignment modes, NP seems to suffer much more than alternative methods, and SZO-1 when applied in per-sample credit often has better alignment than NTWP, something which we did not theoretically explore.
For alignment measurements across all layers in these networks, see Appendix~\ref{app:layerwise-alignment}.

\begin{figure}
    \centering
    \includegraphics[width=\linewidth]{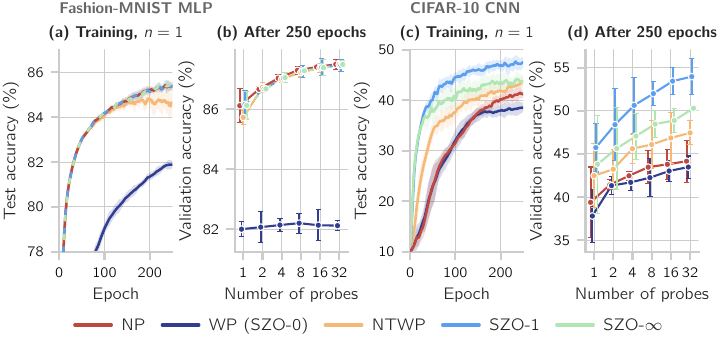}
    \caption{\textbf{The performance of the SZO-family of methods as well as NP, WP, and NTWP when training an MLP (Fashion-MNIST) and a Convolutional Network (CIFAR-10).}
      (a,c) Test accuracy over training with SGD and \(n=1\) (mean over \(10\) seeds, \(\pm 1\) standard deviation); on the MLP, NP, SZO-\(1\) and SZO-\(\infty\) coincide and are drawn interleaved; BP (not shown) reaches \(89.9\%\) and \(83.6\%\).
      (b,d) Validation accuracy after \(250\) epochs against the number of probes per step \(n\) (mean over \(3\) seeds, \(95\%\) confidence intervals; Appendix~\ref{app:probes}).
      The learning rate of each method was selected via a grid search on validation loss.
      Final test accuracies for each method, as well as backpropagation, are reported in Appendix~\ref{app:final-accuracies}.
    }
    \label{fig:performance}
\end{figure}

Figure~\ref{fig:performance} shows that these alignment differences are largely also reflected in the accuracy of models during training. On the MLP, NP and SZO-\(\kappa\) train identically, as expected from their equivalence under per-sample credit. On the CNN the weight-perturbing SZO-\(\kappa\) rules are ahead early and through mid-training; SZO-\(\infty\)'s bias towards a single primary direction narrows its advantage late in training, while SZO-\(1\) stays ahead (Appendix~\ref{app:final-accuracies}). This ordering holds at every number of probes \(n\), and the gap between SZO-\(1\) and NP grows from \(6\) to \(10\) points between \(n=1\) and \(n=32\) (Figure~\ref{fig:performance}d).



\section{Related Work}
Here we place our contributions in the context of ongoing developments in the space of zeroth-order optimisation.

\paragraph{Zeroth-order optimisation and perturbation learning in neural networks.}
Classical ZO methods, including SPSA, finite-difference search, and random gradient-free optimisation, established the basic statistical machinery that perturbation learning relies upon~\citep{spall_multivariate_1992,nesterov_random_2017,flaxman_online_2004,ghadimi_stochastic_2013,shamir_optimal_2015}.
A focus towards neural-network specific applications has additionally brought a broad split into two families: weight perturbation, which injects noise directly into parameters~\citep{jabri_weight_1991,cauwenberghs_fast_1992,werfel_learning_2003}, and node perturbation, which injects noise into activities and assigns credit through the local Jacobian~\citep{barto_neuronlike_1983,mazzoni_more_1991,seung_learning_2003,fiete_gradient_2006,legenstein_reward-modulated_2010}.
\citet{hiratani_stability_2022} recently characterised how both families become unstable as networks deepen, share parameters, and are perturbed simultaneously across layers, and~\citet{zuge_weight_2023} propose a hybrid estimator that empirically combines a weight-level forward pass with a node-level credit assignment (NTWP).
Renewed practical interest in this regime has come from forward-only fine-tuning of large models~\citep{malladi_fine-tuning_2023,zhang_revisiting_2024}, where the attraction is precisely that learning is reduced to inference-like evaluations.
What is missing across this literature is a principled answer to the question of which directions a forward-only method should sample if we assume feature information is available. SZO-\(\kappa\) is the answer to that question, and explains various tradeoffs between NP, WP, and the NTWP methods into a single larger framework.

\paragraph{Structured and guided search directions.}
Several lines of work have further attempted to improve upon vanilla ZO methods by shaping search distributions rather than treating every direction as equally plausible.
Covariance Matrix Adaptation~\citep{hansen_completely_2001} and Evolution Strategies~\citep{salimans_evolution_2017,such_deep_2018} adapt their search covariance from \emph{historical} search paths.
Guided and self-guided ES~\citep{maheswaranathan2019guided,liu_self-guided_2020,ilyas_prior_2018} bias the next probe towards a running estimate of the gradient itself.
Orthogonalised probing~\citep{flugel_beyond_2024,dexheimer_improving_2024} decorrelates random directions, but still treats every parameter direction as a priori equivalent.
\citet{singhal_how_2023} take a distinct approach and form structured gradient guesses by mixing input vectors, local output node sensitivities and even downstream weight matrices to post-hoc improve gradient estimates.
SZO is distinguished from all of these by deriving its search geometry from the \emph{current and local} forward pass: the activation second-moment \(C=AA^\top\) is a free byproduct of inference, and conditioning the search on it requires neither history, an auxiliary gradient estimate or downstream network information, nor explicit orthogonalisation, while sampling a distribution over realisable parameter perturbations whose covariance is shaped by the full second-moment \(C=AA^\top\), with closed-form expected update geometry \(GC^\kappa/\mathrm{Tr}(C^\kappa)\) and controllable rank coverage via \(\kappa\).

\paragraph{Preconditioning and natural-gradient methods.}
Natural-gradient learning~\citep{amari_natural_1998} and its scalable approximations~\citep{martens_optimizing_2015,gupta_shampoo_2018,vyas_soap_2025,jordan2024muon} precondition the \emph{gradient} after it has been computed.
SZO effectively inverts this relationship: it preconditions the \emph{search distribution} so that the resulting estimator already carries a controllable spectral bias \(GC^\kappa/\mathrm{Tr}(C^\kappa)\), and a subsequent whitening step (when desired) either recovers the unbiased gradient or turns the bias into an explicit preconditioned update if gradient structure emerges.
Interestingly, pre-conditioning methods might actually aim to undo the spectral bias which we derive in this work.

\section{Discussion and Limitations}
\paragraph{Empirical takeaways.}
Across a range of settings, SZO-\(\kappa\) methods outperform isotropic WP in alignment with backpropagation, and lead NP when training convolutional networks with SGD.
The largest gains are observed in exactly the regimes the theory predicts: overparameterised fully-connected layers, where \(C\) concentrates rapidly, and convolutional layers, where weight sharing makes independent NP unrealisable in the first place.
All of the proposed SZO methods, as well as NTWP which we show to be equivalent to SZO-1 in certain cases, do far better than traditional WP despite being weight-level perturbation methods proving the efficacy of a tailored search for zeroth-order methods. We therefore recommend SZO-\(1\) where only weights can be perturbed: it beat WP in every setting, by \(3.5\) to \(15\) points, matched NP on fully connected layers and led NP on our CNN with SGD, though not with SOAP. Activation awareness (\(+1.4\) to \(+2.5\) points for \(3\) to \(13\%\) more time; Appendix~\ref{app:activation-awareness}) and, at a fixed compute, more probes on a smaller batch (Appendix~\ref{app:probes}) improve it further.

\paragraph{The cost of spectral shaping.}
Explicitly computing the SZO-\(\kappa\) estimator requires forming \(C^{\kappa/2}\), which seems to require an eigendecomposition of the feature covariance matrix at every step.
For wide modern networks this overhead can dominate the forward pass, and so the most practical versions of SZO are those that preserve the spectral bias without requiring a full matrix decomposition at every update.
In this work, we make use of specific values of $\kappa$ for which this decomposition is cheap to compute.
Specifically, \(\kappa=1\) and \(\kappa=\infty\) both admit decomposition-free implementations for linear and convolutional layers (Appendix~\ref{iswp_details}).
\(\kappa=2\) is a third natural choice that needs only matrix products, and our sweep (Appendix~\ref{app:kappa-sweep}) shows that alignment has already saturated by \(\kappa=2\), although we do not train with it here.
NTWP provides a further route: with per-batch credit it achieves the SZO-\(1\)-style spectral shaping through the activity-derived eligibility, without ever explicitly forming \(C^{1/2}\).
Thus, although the full SZO-\(\kappa\) family has a nontrivial spectral cost in its most literal form, a range of useful choices we consider here can be implemented without expensive decompositions.

\paragraph{Deep network crosstalk.}
When all layers are perturbed simultaneously, every estimator considered here suffers from inter-layer `crosstalk': the loss change caused by other layers' noise enters each layer's estimate.
As shown in Appendix~\ref{app:inter-layer-variance}, this crosstalk cancels in expectation but adds variance terms that grow with the number of simultaneously perturbed layers.
The cleanest mitigation is iterative layer-wise perturbation, which we made use of for all main figures in this paper, at the cost of \(\mathcal{O}(L)\) forward passes for a network of depth \(L\) (Appendix~\ref{app:corrections}).
This is a real compute tradeoff, but not an absolute requirement. In Appendix~\ref{app:seq-vs-sim} we find that simultaneously perturbing all layers remains effective on the MLP and for the SZO-\(\kappa\) estimators on the CNN, albeit with a penalty in peak performance.
This suggests that crosstalk can be treated as a variance-compute tradeoff rather than a fundamental obstruction to such methods, and developing constant-time variance-reduction strategies for specific zeroth-order applications is a natural next step.

\section{Conclusion}
Zeroth-order optimisation has long suffered from the curse of dimensionality, with random search directions becoming increasingly misaligned with the true gradient as parameter count grows.
The central observation of this work is that the forward pass already exposes useful geometry: a layer's input second moment records, at no extra cost, which parameter directions can causally affect the network on the current data.
Building on this observation we derived the SZO-\(\kappa\) family of spectral zeroth-order estimators, and showed that it unifies the existing perturbation literature under a single parameter: \(\kappa=0\) recovers weight perturbation, any non-zero choice of \(\kappa\) recovers node perturbation in fully connected layers with per-sample credit, and \(\kappa=1\) matches the hybrid perturbation method of~\citet{zuge_weight_2023} in expectation under per-batch credit, which we formalise here as Node-Through-Weight Perturbation.
On Fashion-MNIST and CIFAR-10 these estimators substantially improve gradient alignment over vanilla zeroth-order optimisation methods and show successful learning.
We see SZO-\(\kappa\), together with their realisability on shared-parameter architectures, as a step towards forward-only learning on hardware where backpropagation is undesirable or unavailable.

\begin{ack}
This publication is part of the project Dutch Brain Interface Initiative (DBI$^2$) with project number 024.005.022 of the research programme Gravitation which is (partly) financed by the Dutch Research Council (NWO).

This work used the Dutch national e-infrastructure with the support of the SURF Cooperative using grant no. EINF-18863.
\end{ack}

\bibliographystyle{plainnat}
\bibliography{./references}


\clearpage
\appendix
\setcounter{table}{0}
\renewcommand{\thetable}{A.\arabic{table}}
\renewcommand{\theHtable}{appendix.\arabic{table}}
\let\appendixsection\section
\renewcommand{\section}{\FloatBarrier\appendixsection}

\section{Gradient Guessing Based Upon the Local Response}
\label{app:response-surrogates}

Suppose one wishes to choose an ideal perturbation for a given layer before they have any knowledge of the downstream gradient.
We can formalise this such that a layer, \(H=WA\), with inputs \(A\in\mathbb{R}^{q\times B}\), when combined with a perturbation \(\Psi\in\mathbb{R}^{k\times q}\) produces \(\Delta H=\Psi A\).
Writing the downstream error as \(V=\nabla_H\mathcal{L}_B\), the first-order loss response is
\[
  \Delta\mathcal{L}_B
  =
  \epsilon\langle V,\Psi A\rangle_F+O(\epsilon^2)
  =
  \epsilon\langle VA^\top,\Psi\rangle_F+O(\epsilon^2).
\]
If \(V\) were known, the best fixed-norm perturbation would align with \(VA^\top\), which is exactly the unknown gradient.
The useful question is therefore: what should be done when the forward pass gives \(A\), but gives no directional information about \(V\)?

The least aggressive assumption is that downstream errors are rotationally invariant across the activity coordinates and independent of the input-feature second moment.
For example, assume entries of \(V\) have zero mean and satisfy
\[
  \mathbb{E}_V[V^\top V]=\alpha I_B
  \quad\text{or, equivalently for this calculation,}\quad
  \mathbb{E}_V[\langle V,B_1\rangle_F\langle V,B_2\rangle_F]
  =
  \alpha\langle B_1,B_2\rangle_F .
\]
Then the expected squared response of a candidate perturbation is
\[
  \mathbb{E}_V[\langle V,\Psi A\rangle_F^2]
  =
  \alpha\|\Psi A\|_F^2
  =
  \alpha\,\mathrm{Tr}(\Psi C\Psi^\top),
  \qquad C=AA^\top .
\]
Thus, when downstream-gradient covariance is unknown and isotropy is the simplest assumption, the activation second-moment matrix \(C\) is the only available geometry by which we can bias our search toward the true gradient.
In this sense, SZO methods solve a local surrogate gradient alignment objective using the known feature geometry and an isotropic prior over the unknown downstream error.

\paragraph{When is the isotropy prior actually justified?}
The prior has two justifications. At initialisation, standard init schemes (orthogonal, He, dynamical-isometry) make \(V\) rotationally symmetric in distribution across activity coordinates over the weight ensemble; this is the basis of NTK and signal-propagation analyses~\citep{saxe_exact_2014,schoenholz_deep_2017,jacot_neural_2020,pennington_emergence_2018}.
At any point in training, the isotropic prior is also the least informative choice when no information beyond the forward pass is used: it assumes no preferred direction for \(V\) and no dependence of \(V\) on \(C\). It is not guaranteed to hold once training introduces structure into \(V\), in which case the response surrogate is a principled heuristic rather than an exact optimum.

\section{Rank Collapse, Nuclear-Norm Coverage, and the Matched-Filter Derivation}
\label{app:matched-filter}

Appendix~\ref{app:response-surrogates} shows that, when the downstream gradient covariance is unknown, an isotropic prior over downstream errors reduces the local problem to the known feature second moment \(C=AA^\top\).
That result identifies a geometry, but it does not yet identify the right \emph{utility} for learning.
If the utility is simply ``make the scalar loss response as large as possible,'' then the optimal perturbation is allowed to spend its entire budget on whichever feature direction gives the largest immediate response.
This is the estimator which we identify as SZO-\(\infty\).
This is attractive when the feature spectrum is sharply concentrated, when very few probes are available, or when one deliberately wants the strongest activity response per unit perturbation energy.
The cost, however, of such a choice is that directions outside the leading eigenspace are not explored at all and we only have a rank-1 weight update for each probe explored.

We explore this outcome as well as what can be done to move beyond such a restricted estimator in this appendix.
First, we show that the maximum-response objective gives a legitimate rank-1 estimator, SZO-\(\infty\).
Second, we show that one can choose a different utility, namely rank-covering response rather than maximum scalar response, which leads to the matched-filter perturbation used by SZO-\(1\).
The response-energy surrogate from Appendix~\ref{app:response-surrogates} gives
\[
  \max_{\|\Psi\|_F^2=c}\mathrm{Tr}(\Psi C\Psi^\top).
\]

To see the solution directly, write the eigendecomposition \(C=U\Lambda U^\top\), with eigenvectors \(u_j\) and eigenvalues \(\lambda_1\ge\lambda_2\ge\cdots\ge 0\).
Decompose each row of \(\Psi\) in this eigenbasis.
Equivalently, let \(z_j=\Psi u_j\in\mathbb{R}^{k}\) denote the output-side perturbation attached to input eigenvector \(u_j\).
Then
\[
  \mathrm{Tr}(\Psi C\Psi^\top)
  =
  \sum_j \lambda_j\|z_j\|_2^2,
  \qquad
  \sum_j\|z_j\|_2^2=c.
\]
The optimisation should now be clear: \(\|z_j\|_2^2\) is the perturbation energy placed on feature direction \(u_j\), and each unit of energy returns \(\lambda_j\) units of expected response.
The maximum therefore assigns all energy to the largest eigenvalue:
\[
  z_1=\sqrt{c}\,r,\qquad z_j=0\;\;(j>1),
  \qquad \|r\|_2=1,
\]
which gives
\[
  \Psi^\star_{\infty}=\sqrt{c}\,r u_1^\top .
\]
Randomising \(r\) isotropically across output units gives the stochastic SZO-\(\infty\) estimator.
It is optimal for maximum expected squared response under the isotropic downstream-gradient prior.
However, its expected update keeps only the leading feature component,
\[
  \mathbb{E}[\widehat{G}_{\infty}]
  \propto
  (G u_1)u_1^\top,
\]
which explains both its strength and its limitation: it has low variance and high response, but it discards gradient components outside the leading eigenspace.

SZO-\(\infty\) is therefore one extreme of the space of options for perturbation.
To obtain coverage, we do not have to change the underlying energy quantity, but to evaluate the same induced-activity energy under a concave penalty on its singular spectrum.
Replacing each \(\mathrm{Tr}(\Psi C\Psi^\top) = \sum_i s_i(\Psi A)^2\) by a sum over the square roots yields the nuclear norm
\[
  \|\Psi A\|_*
  =
  \sum_i s_i(\Psi A)
  =
  \mathrm{Tr}\!\left((\Psi C\Psi^\top)^{1/2}\right).
\]
Under the same perturbation budget, this values every induced singular direction linearly instead of quadratically, encouraging rank coverage.

The corresponding constrained problem is
\[
  \max_{\|\Psi\|_F^2=c}
  \mathrm{Tr}\!\left((\Psi C\Psi^\top)^{1/2}\right).
\]
Using the matrix derivative
\[
  \nabla_\Psi\,\mathrm{Tr}\!\left((\Psi C\Psi^\top)^{1/2}\right)
  =
  (\Psi C\Psi^\top)^{-1/2}\Psi C,
\]
we enforce the budget with a multiplier \(\gamma\) and form the Lagrangian
\[
  \mathcal{J}(\Psi,\gamma)
  =
  \mathrm{Tr}\!\left((\Psi C\Psi^\top)^{1/2}\right)
  -\frac{\gamma}{2}\left(\|\Psi\|_F^2-c\right).
\]
Since \(\nabla_\Psi\|\Psi\|_F^2=2\Psi\), setting \(\nabla_\Psi\mathcal{J}=0\) gives the stationarity condition
\[
  (\Psi C\Psi^\top)^{-1/2}\Psi C-\gamma\Psi=0
  \quad\Longleftrightarrow\quad
  (\Psi C\Psi^\top)^{-1/2}\Psi C=\gamma\Psi .
\]
A matched-filter solution \(\Psi=\beta RC^{1/2}\) satisfies this condition whenever \(R\) has orthonormal columns, \(R^\top R=I\), which requires \(k\ge q\) (for \(k<q\) it holds only approximately); the output-side directions of \(R\) are otherwise completely free.
To check this by substitution, \(R^\top R=I\) gives \(\Psi C\Psi^\top=\beta^2RC^2R^\top=\beta^2(RCR^\top)^2\), so \((\Psi C\Psi^\top)^{-1/2}=\tfrac{1}{\beta}RC^{-1}R^\top\) (a pseudo-inverse when \(k>q\)), and
\[
  (\Psi C\Psi^\top)^{-1/2}\Psi C
  =
  \tfrac{1}{\beta}RC^{-1}R^\top\cdot\beta RC^{3/2}
  =
  RC^{1/2}
  =
  \tfrac{1}{\beta}\Psi ,
\]
so the condition holds with \(\gamma=1/\beta\).
As for SZO-\(\infty\), SZO-\(1\) samples from this family of optima by drawing \(R\) isotropically with entry variance \(1/k\): \(\mathbb{E}[R^\top R]=I_q\), so \(\mathbb{E}\|\Psi\|_F^2=\beta^2\mathrm{Tr}(C)=c\) as below, so the spectral shaping \(C^{1/2}\) is exact in expectation and only the output-side directions are random.
This solution equalises the marginal expected response per searched direction: each unit of perturbation energy returns the same nuclear-norm contribution.
The norm constraint fixes
\[
  \|\beta RC^{1/2}\|_F^2
  =
  \beta^2\mathrm{Tr}(C)=c,
  \qquad
  \beta=\sqrt{\frac{c}{\mathrm{Tr}(C)}} .
\]
Therefore the rank-covering optimum has the form
\[
  \Psi^\star
  =
  \sqrt{\frac{c}{\mathrm{Tr}(C)}}\,RC^{1/2}.
\]
The rank-1 solution (\(\kappa\to\infty\)), the matched filter (\(\kappa=1\)) and isotropic weight perturbation (\(\kappa=0\)) are all members of a single family obtained by raising \(C\) to a power,
\[
  \Psi_\kappa
  =
  \sqrt{\frac{c}{\mathrm{Tr}(C^\kappa)}}\,RC^{\kappa/2}.
\]
This is a natural way to interpolate between these solutions: rescaling the inputs multiplies every power of \(C\) by a single constant, which the normaliser removes, so the relative emphasis on each feature direction does not depend on the scale of the inputs.
The parameter \(\kappa\) controls the spectral tradeoff: \(\kappa=0\) is isotropic, \(\kappa=1\) is matched filtering, larger \(\kappa\) increasingly emphasises dominant features, and \(\kappa\to\infty\) recovers rank collapse.

\begin{algorithm}[t]
\caption{SZO-\(\kappa\) for one layer}
\label{alg:szo-main}
\small
\begin{algorithmic}[1]
\State \textbf{Input:} layer \(W\), batch activations \(A\), loss \(\mathcal{L}_B\), scale \(\epsilon\), exponent \(\kappa\), draws \(n\), budget \(c\)
\State Run a clean forward pass; set \(H\gets WA\), \(C\gets AA^\top\), and baseline loss \(\mathcal{L}_0\gets\mathcal{L}_B(W)\)
\State Initialise \(\widehat{G}\gets 0\)
\For{\(i=1,\ldots,n\)}
  \State Draw mean-zero isotropic \(R_i\in\mathbb{R}^{k\times q}\) (iid entries of variance \(1/k\))
  \State Form \(\Psi_{\kappa,i}\gets \sqrt{c/\mathrm{Tr}(C^\kappa)}\,R_i C^{\kappa/2}\)
  \State Evaluate \(\mathcal{L}_i^+\gets\mathcal{L}_B(W+\epsilon\Psi_{\kappa,i})\)
  \State Accumulate \(\widehat{G}\gets \widehat{G}+(\mathcal{L}_i^+-\mathcal{L}_0)\Psi_{\kappa,i}/(n\epsilon c)\)
\EndFor
\State \Return \(\widehat{G}\)
\end{algorithmic}
\end{algorithm}
Two-sided evaluations replace the one-sided loss difference exactly as in SPSA, and in practice the fixed budget \(c\) may be replaced by the realised \(\|\Psi_{\kappa,i}\|_F^2\) for per-draw normalisation.

\section{Equivalences Between WP, NP, SZO-\texorpdfstring{\(\kappa\)}{kappa}, and NTWP}
\label{app:equivalences}

\paragraph{SZO-\(0\) recovers WP.}
For \(\kappa=0\), \(C^0=I\), and
\[
  \Psi_0=\sqrt{\frac{c}{\mathrm{Tr}(I)}}R.
\]
Up to the scalar normalisation, this is ordinary isotropic parameter noise.
Correlating the loss change with \(\Psi_0\) is therefore weight perturbation or one-sided SPSA.
Here \(C^0=I\) assumes that the zero eigenvalues of \(C\) are not ignored, and under it SZO-\(0\) is WP for every \(C\); this is how WP and the \(\kappa=0\) points of Appendix~\ref{app:kappa-sweep} are run.
The limit \(\kappa\to0^{+}\) only agrees with this assumption when \(C\) has full rank: otherwise \(C^{\kappa/2}\) is low rank and tends to an orthogonal projector \(P\) onto the span of the inputs, and \(\mathrm{Tr}(C^\kappa)\) to \(\mathrm{rank}(C)\), so the noise is isotropic within that span and zero outside it.
Since \(G=VA^\top\) lies in that span (\(GP=G\)), the limit is also unbiased and only drops noise directions that cannot change the layer's output on the batch.
Full rank is typical for a per-batch \(C\) of a convolution, whose many patches span its input, but a per-sample \(C_b=a_ba_b^\top\) of a fully connected layer has rank one, so there the limit is NP rather than WP, like every \(\kappa>0\) (below).

\paragraph{Sample-wise SZO-\(\infty\) recovers NP in fully connected layers.}
For a single fully connected example \(a\), \(C=aa^\top\) is rank one with \(u=a/\|a\|_2\).
In the \(\kappa\to\infty\) limit,
\[
  \Psi_\infty=\sqrt{c}\,(Ru)u^\top= \sqrt{c}\,\phi u^\top,
\]
where \(\phi=Ru\) is isotropic in output/activity space when \(R\) has isotropic rows.
The induced activity perturbation is
\[
  \Delta h=\Psi_\infty a=\sqrt{c}\,\|a\|_2\,\phi ,
\]
which is exactly node perturbation up to scale.
The corresponding weight eligibility is
\[
  \phi a^\top=J_{h,W}^\top\phi ,
\]
the usual NP eligibility for \(h=Wa\).
For a mini-batch with sample-wise losses, applying this construction independently to each \(a_b\) and averaging recovers batch NP.

\paragraph{Shared-parameter layers break exact NP equivalence.}
In convolutional layers, the columns of \(A\) are receptive-field patches over examples and spatial positions.
Even for one image, \(C=AA^\top\) is usually high-rank.
A single shared kernel perturbation induces correlated node perturbations across positions:
\[
  \Delta h_p=\Delta K\,a_p .
\]
Independent NP noise at every position would require a different \(\Delta K_p\) for each patch, violating weight sharing.
Thus SZO-\(\infty\) is the realisable shared-parameter analogue of NP, not identical independent node perturbation.

\paragraph{NTWP recovers the SZO-\(1\) expected geometry.}
The equivalence is easiest to see by deriving the expected update from both sides.
Let
\[
  G=\nabla_W\mathcal{L}_B=VA^\top,
  \qquad C=AA^\top .
\]
First consider explicit SZO-\(1\).
It draws a spectrally shaped weight perturbation
\[
  \Psi_1=RC^{1/2},
\]
omitting scalar normalisation.
The first-order loss change is
\[
  \Delta\mathcal{L}_{\mathrm{SZO\mbox{-}1}}
  \approx
  \epsilon\langle G,RC^{1/2}\rangle_F .
\]
The SZO-\(1\) update correlates this loss change with the same shaped direction \(RC^{1/2}\), so
\[
  \mathbb{E}_R[
  \langle G,RC^{1/2}\rangle_F\,RC^{1/2}]
  =
  GC,
\]
where the final equality follows from isotropy of \(R\).
Thus SZO-\(1\) produces the expected geometry \(GC\), up to normalisation by \(\mathrm{Tr}(C)\) and finite-difference error.

Now consider NTWP.
For \(H=WA\), draw isotropic parameter noise \(R\) without spectral shaping.
It induces the activity perturbation
\[
  \Delta H=RA.
\]
Mapping this induced activity perturbation back to weights gives the local eligibility
\[
  E=\Delta H A^\top=RAA^\top=RC.
\]
The first-order loss change is measured through the original parameter perturbation:
\[
  \Delta\mathcal{L}_{\mathrm{NTWP}}
  \approx
  \epsilon\langle G,R\rangle_F .
\]
NTWP then correlates this scalar with the activity-derived eligibility \(RC\):
\[
  \mathbb{E}_R[
  \langle G,R\rangle_F\,RC]
  =
  GC.
\]
The two estimators therefore arrive at the same first-order expected geometry by different routes:
\[
  \underbrace{\langle G,RC^{1/2}\rangle_F\,RC^{1/2}}_{\mathrm{SZO\mbox{-}1}}
  \quad\Longrightarrow\quad
  GC,
  \qquad
  \underbrace{\langle G,R\rangle_F\,RC}_{\mathrm{NTWP}}
  \quad\Longrightarrow\quad
  GC.
\]
SZO-\(1\) puts the spectral factor into the perturbation itself, whereas NTWP puts the spectral factor into the eligibility generated by the induced activity change.
This is why NTWP implements the matched-filter bias without explicitly forming \(C^{1/2}\).

\paragraph{Per-sample credit breaks the NTWP/SZO-\(1\) equivalence.}
The derivation above correlates one scalar loss difference for the whole batch with the eligibility \(RC\).
With per-sample credit, example \(b\) instead correlates its own loss difference \(\Delta\mathcal{L}_b\approx\epsilon\,g_b^\top Ra_b\), with \(g_b=\nabla_{h_b}\mathcal{L}_b\), with its own induced activity perturbation \(\Delta h_b=\epsilon Ra_b\), normalised by its energy \(\|\Delta h_b\|_2^2/k\).
Writing \(v_b=Ra_b\), which is isotropic so that \(\mathbb{E}[v_bv_b^\top/\|v_b\|_2^2]=I_k/k\),
\[
  \mathbb{E}\Big[\frac{1}{B}\sum_b k\,\frac{(g_b^\top v_b)\,v_b}{\|v_b\|_2^2}\,a_b^\top\Big]
  =
  \frac{1}{B}\sum_b g_b a_b^\top
  =
  G .
\]
Per-sample NTWP is therefore an unbiased estimator of \(G\) without any spectral factor: in expectation it is node perturbation, with node noise induced by a single realisable weight perturbation shared by the batch.
NTWP and SZO-\(1\) share the geometry \(GC\) only under per-batch credit.
(With the per-example normaliser above, per-batch NTWP gives \(G\,\tfrac{1}{B}\sum_b\hat{a}_b\hat{a}_b^\top\) with \(\hat{a}_b=a_b/\|a_b\|_2\), which matches \(GC\) up to scale when the inputs have equal norms.)
All NTWP training results in this work use per-sample credit.

\begin{algorithm}[t]
\caption{Node-Through-Weight Perturbation for one layer (per-batch credit)}
\label{alg:ntwp-main}
\small
\begin{algorithmic}[1]
\State \textbf{Input:} layer \(W\), batch activations \(A\), loss \(\mathcal{L}_B\), scale \(\epsilon\), draws \(n\)
\State Run a clean forward pass; set \(H\gets WA\), \(C\gets AA^\top\), and baseline loss \(\mathcal{L}_0\gets\mathcal{L}_B(W)\)
\State Initialise \(\widehat{G}\gets 0\)
\For{\(i=1,\ldots,n\)}
  \State Draw mean-zero isotropic parameter noise \(R_i\in\mathbb{R}^{k\times q}\)
  \State Evaluate \(\mathcal{L}_i^+\gets\mathcal{L}_B(W+\epsilon R_i)\)
  \State Compute induced activity perturbation \(\Delta H_i\gets R_iA\)
  \State Map activity credit to weights \(E_i\gets \Delta H_iA^\top=R_iC\)
  \State Normalise \(c_i\gets \|\Delta H_i\|_F^2=\mathrm{Tr}(R_iCR_i^\top)\)
  \State Accumulate \(\widehat{G}\gets \widehat{G}+(\mathcal{L}_i^+-\mathcal{L}_0)E_i/(n\epsilon c_i)\)
\EndFor
\State \Return \(\widehat{G}\)
\end{algorithmic}
\end{algorithm}

\section{Optional Gradient Corrections}
\label{app:corrections}

We collect here some optional stabilising methods from the literature that can be layered on top of any zeroth-order method, each targeting a distinct source of error in the generic linearised estimator
\(
  \hat{g}_n
  =
  g M_n + O(\epsilon^2) + \xi ,
\)
where \(M_n\) is the effective search covariance, \(O(\epsilon^2)\) is finite-difference truncation error, and \(\xi\) represents errors induced by using finite numbers of probes and inter-layer noise.
Estimator-specific \(M_n\)'s are listed in Appendix~\ref{app:bias-variance-tradeoffs}, and full derivations are deferred to Appendices~\ref{app:whitening-corrections} through \ref{app:bias-variance-tradeoffs}.
Table~\ref{tab:corrections} summarises these corrections, the error term each addresses, and their original references.

\begin{table}[h]
\centering
\small
\caption{Optional corrections that can be layered on top of zeroth-order and perturbation-based gradient estimators. Each targets a distinct source of error in \(
  \hat{g}_n
  =
  g M_n + O(\epsilon^2) + \xi ,
\).}
\label{tab:corrections}
\begin{tabular}{p{0.20\linewidth} p{0.55\linewidth} p{0.17\linewidth}}
\toprule
\textbf{Correction} & \textbf{What it does} & \textbf{Origin} \\
\midrule
Iterative layer-wise perturbation &
Sequentially perturbs a single layer at a time across a network, removing inter-layer crosstalk terms for an extra \(L\times\) forward-pass cost. When all layers are instead perturbed simultaneously, the cross-layer covariance vanishes in expectation for all methods, but loss changes driven by other layers' noise inflate each layer's within-layer variance (Appendix~\ref{app:inter-layer-variance}). &
\citet{dalm_effective_2023} \\
\midrule
Activation Awareness &
When perturbing nodes, or considering the effect of weight perturbation at the node level, one can additionally take into account the locally-available derivative of the node's activation function. This ultimately can allow ensuring that the updates applied to weights are guided by how susceptible the output nodes are to the weight change (Appendix~\ref{app:activation-awareness}).
 &
\citet{singhal_how_2023} \\

\midrule
Whitening / preconditioning &
Absorbs the rotation \(\mathbb{E}[\hat{g}_n]\approx M_{\mathrm{out}}gM_{\mathrm{in}}\) induced by \(M_n\neq I\) using diagonal Adam-style second moments, Kronecker-factored Shampoo/SOAP, or Newton--Schulz orthogonalisation as in Muon. For SZO-\(\kappa\) this turns the intended spectral bias into an explicit preconditioned step; the same correction applies to NTWP, whose activity-derived eligibility shares SZO-\(1\) geometry. &
\citet{jordan2024muon, lang_powering_2026}; see also~\citet{kinmga2015method,gupta_shampoo_2018,vyas_soap_2025} \\
\bottomrule
\end{tabular}
\end{table}

\section{Whitening Corrections for Perturbation Estimators}
\label{app:whitening-corrections}

Let \(e_i\) denote the eligibility used by a perturbation estimator after the scalar loss difference has been measured.
For a one-sided estimator at scale \(\epsilon\),
\[
  \widehat{g}
  =
  \frac{1}{n\epsilon}\sum_{i=1}^n
  \big(\mathcal{L}(\theta+\epsilon\psi_i)-\mathcal{L}(\theta)\big)e_i .
\]
In the local linear regime,
\[
  \mathcal{L}(\theta+\epsilon\psi_i)-\mathcal{L}(\theta)
  =
  \epsilon\,g\psi_i+O(\epsilon^2),
\]
so the expected update is controlled by the cross-covariance between the measured perturbation \(\psi_i\) and the eligibility \(e_i\):
\begin{equation}
  \mathbb{E}[\widehat{g}]
  =
  g\,\mathbb{E}[\psi_i e_i^\top]+O(\epsilon^2)
  =
  gM+O(\epsilon^2).
  \label{eq:app-generic-bias}
\end{equation}
For WP and SPSA, \(e_i=\psi_i/\sigma^2\), hence \(M=I\) in population.
For finite \(n\), however, the realised matrix \(\widehat M=(n\sigma^2)^{-1}\sum_i\psi_i\psi_i^\top\) can be far from \(I\), producing the finite-batch covariance rotation described in the main text.
For SZO-\(\kappa\), with \(\Psi_i=R_iC^{\kappa/2}\) and isotropic \(R_i\),
\[
  \mathbb{E}[\widehat{G}_\kappa]
  \propto
  G C^\kappa ,
\]
up to the scalar normalisation used in Algorithm~\ref{alg:szo-main}.
For NTWP, \(e_i=R_iC\) while the loss is measured with \(R_i\), yielding the same first-order geometry \(GC\).

When the layer update is matrix-valued, many of these covariances factor on the output and input sides:
\begin{equation}
  \mathbb{E}[\widehat{G}]
  \approx
  M_{\mathrm{out}}G M_{\mathrm{in}}.
  \label{eq:app-factor-bias}
\end{equation}
The damped whitening correction is therefore
\begin{equation}
  \widehat{G}^{\mathrm{white}}
  =
  (M_{\mathrm{out}}+\lambda I)^{-1}
  \widehat{G}
  (M_{\mathrm{in}}+\lambda I)^{-1}.
  \label{eq:app-whitened-grad}
\end{equation}
If the goal is an unbiased Euclidean gradient estimate, Eq.~\eqref{eq:app-whitened-grad} removes the covariance factors in Eq.~\eqref{eq:app-factor-bias}.
If the goal is faster optimisation, the same operation can instead be interpreted as a layer-wise natural-gradient or KFAC-style preconditioned step: \(M_{\mathrm{in}}\) is estimated from input activations and \(M_{\mathrm{out}}\) from output-side sensitivity statistics.

\begin{table}[h]
\centering
\footnotesize
\caption{\textbf{Whitening corrections.}}
\label{tab:whitening}
\begin{tabular}{@{}p{0.13\textwidth}p{0.35\textwidth}p{0.30\textwidth}p{0.13\textwidth}@{}}
\toprule
Method & Effective covariance & Whitening correction & Practical form \\
\midrule
WP/SPSA &
\(I\) in population; realised \(\widehat M-I\) at finite \(n\) &
optional correction by \(\widehat M^{-1}\) or optimiser state &
Adam, Muon \\
NP &
activity-noise covariance \(M_\phi\) &
whiten active activity locus before mapping to weights &
Adam, Muon \\
SZO-\(\kappa\) &
input factor \(C^\kappa/\mathrm{Tr}(C^\kappa)\) &
right multiply by \((C^\kappa+\lambda I)^{-1}\), or keep as preconditioner &
Shampoo/SOAP \\
NTWP &
activity-induced factor \(C\) &
right whitening of activity eligibility; optional two-sided correction &
preconditioned \\
\bottomrule
\end{tabular}
\end{table}

The damping \(\lambda\) is essential in low-rank or saturated regimes.
Whitening exactly along directions with near-zero covariance would amplify noise in directions that were barely sampled.
In practice one either damps, truncates small eigenvalues, or whitens only on the active subspace identified by the current batch.

\section{Inter-Layer Variance and Crosstalk}
\label{app:inter-layer-variance}

Consider a depth-\(L\) network in which all layers are perturbed simultaneously.
Let \(P_l\) be the stochastic object used at layer \(l\): a weight perturbation for WP/SZO, an activity perturbation for NP, or, for NTWP, a weight perturbation \(R_l\) whose own induced activity perturbation \(R_lA_{l-1}\) forms the eligibility.
The first-order loss change decomposes as
\begin{equation}
  \Delta\mathcal{L}
  \approx
  \epsilon\sum_{p=1}^L\langle G_p,P_p\rangle .
  \label{eq:app-layer-loss-sum}
\end{equation}
The estimator for layer \(l\) has the form \(\widehat G_l\propto\Delta\mathcal{L}\,E_l\), where \(E_l\) is the layer eligibility trace.
The second moment between layers \(k\) and \(l\) is
\[
  M_{kl}
  =
  \mathbb{E}[\widehat G_k\otimes\widehat G_l].
\]
When \(k\neq l\), independent mean-zero layer perturbations make the only surviving pairings factor into
\[
  M_{kl}
  =
  \mathbb{E}[\widehat G_k]\otimes\mathbb{E}[\widehat G_l]
\]
for WP, NP, SZO, and NTWP.
Thus their spurious cross-layer covariance vanishes in expectation.
The within-layer case is different because the outer factor \(E_l\otimes E_l\) is even in the layer-\(l\) noise.
Squaring Eq.~\eqref{eq:app-layer-loss-sum} gives a double sum over layer pairs \((p,p')\); its diagonal terms with \(p=p'\neq l\) survive and add a positive variance contribution from every other layer:
\begin{equation}
  \operatorname{Var}(\widehat G_l)
  \approx
  \operatorname{Var}_{\mathrm{isolated}}(\widehat G_l)
  +
  \Big(\sum_{p\neq l}\|G_p\|_F^2\Big)S_l .
  \label{eq:app-variance-inflation}
\end{equation}
The shape \(S_l\) depends on the eligibility: for WP it is approximately isotropic in parameter space; for NP it is activity-noise covariance mapped through the local input; for SZO it carries the feature geometry induced by \(C_l^\kappa\); and for NTWP it is the covariance of the layer's own induced activity perturbation mapped back through the local input, which is \(C_l\)-shaped like SZO-\(1\) under per-batch credit and NP-like under per-sample credit.

NTWP falls in the same class because its eligibility \(E_l=R_lA_{l-1}A_{l-1}^\top\) is built from the layer's own injected perturbation only.
Under simultaneous perturbation the layer input is the noisy \(\tilde A_{l-1}\), but the upstream contribution to \(R_l\tilde A_{l-1}\) is second order in \(\epsilon\) and has zero mean, since \(R_l\) appears in it only once.

The situation differs if the eligibility is instead built from the \emph{measured} activity change (noisy minus clean output), as in activity-based node perturbation~\citep{dalm_effective_2023}.

\section{Activation-Aware Credit Assignment}
\label{app:activation-awareness}

Consider a perturbed layer with pre-activations \(h\in\mathbb{R}^k\) followed by an elementwise nonlinearity \(y=\rho(h)\).
By the chain rule the pre-activation gradient is
\[
  \nabla_h\mathcal{L}
  =
  s\odot\nabla_y\mathcal{L},
  \qquad
  s=\rho'(h),
\]
where \(s\) is evaluated at the clean forward pass.
A unit with \(s_i\approx 0\), such as an inactive ReLU, has no first-order effect on the loss: any credit assigned to it adds variance to the estimate without adding signal.
Like the activation second moment \(C=AA^\top\), the slope \(s\) is available locally during the forward pass at no additional cost, and can therefore be used to shape credit assignment without any backward computation.

\paragraph{The awareness weight \(w=s^2\).}
We weight the credit assigned to each unit by \(w_i\) and choose \(w\) using the same isotropic-prior argument as in Appendix~\ref{app:response-surrogates}, now placed on the post-activation gradient.
Let \(\widehat{g}_i\) be an unbiased per-unit estimate of \((\nabla_h\mathcal{L})_i\) with approximately unit-independent variance \(\nu\), and take the update \(d_i=w_i\widehat{g}_i\).
Under an iid\ prior \(\mathbb{E}[(\nabla_y\mathcal{L})_i^2]=\alpha\), a step of size \(\eta\) yields the expected first-order decrease and second-order variance cost
\[
  \eta\,\alpha\sum_i w_i s_i^2
  \qquad\text{and}\qquad
  \tfrac{\eta^2}{2}\,\beta\,\nu\sum_i w_i^2 ,
\]
respectively, for a local curvature scale \(\beta\).
Maximising their difference over \(w\) gives \(w_i\propto s_i^2\).
This weight has two useful properties.
First, it is non-negative for any activation function, so non-monotone activations (e.g.\ GELU, SiLU) never flip the sign of a unit's credit.
Second, for ReLU and max-pooling it is an exact \(0/1\) mask (for the leaky ReLU of our networks, \(w\in\{10^{-4},1\}\), approximately so), so the expected update on the active units is unchanged and the variance contributed by inactive units is removed entirely.

\paragraph{Node-space estimators (NP, NTWP).}
For estimators whose eligibility lives in activity space, the node credit \(\phi\) is replaced by \(\phi\odot w\) before it is mapped back to weights through the local input,
\[
  \widehat{\nabla_W\mathcal{L}}
  \propto
  \sum_{i=1}^n \Delta\mathcal{L}_i\,(\phi_i\odot w)\,a^\top ,
\]
while the normaliser is kept as the unweighted perturbation energy.
For NTWP, \(\phi_i\) is the induced activity perturbation \(\Delta h_i=R_ia\).
For a \(0/1\) slope the expected node credit is therefore exactly the node gradient on the active units.

\paragraph{Weight-space estimators (WP, SZO-\(1\), SZO-\(\infty\)).}
For estimators that correlate the loss change with a weight perturbation \(\Psi\), activation awareness only modifies the readout, not the perturbation itself.
Each loss difference is weighted per example \(b\) and output channel \(j\) before the batch is aggregated,
\[
  \widehat{G}_{j,:}
  \propto
  \sum_{b}\Delta\mathcal{L}_b\,\tilde{w}_{bj}\,\Psi_{j,:},
\]
so that the update to row \(j\) of \(W\) only receives credit from examples in which output \(j\) can affect the loss.
For a linear layer \(\tilde{w}_{bj}=w_{bj}\).
For a convolutional layer the slope varies over spatial positions \(\ell\), and is averaged with weights \(p_{bj\ell}\) given by how strongly the probe changed each position's output,
\[
  \tilde{w}_{bj}
  =
  \frac{\sum_\ell w_{bj\ell}\,p_{bj\ell}}{\sum_\ell p_{bj\ell}},
\]
with \(p_{bj\ell}=(\Delta y_{bj\ell})^2\) in general and \(p_{b\ell}=\langle \hat{a}_b,x_{b,\ell}\rangle^2\) for the energy-weighted SZO-\(\infty\) construction of Appendix~\ref{iswp_details}.
The weights are used without renormalisation.
On a linear layer a \(0/1\) slope therefore masks whole rows, and each estimator retains its own expected update on the active rows.
On a convolutional layer the averaged weight scales a channel's update rather than masking positions: behind a \(2\times2\) max-pool only about one position in four passes gradient, so the step of that layer shrinks by about the same factor.
In all cases, \(w\equiv 1\) recovers the activation-unaware estimator exactly.

\paragraph{Scope.}
The slope must come from a local, per-unit map between the perturbed layer and its successor: elementwise activations, affine maps, residual sums (holding the other summand at its clean value), and non-overlapping max-pooling.
Normalisation layers and dropout couple units or examples, so \(s\) is no longer a per-unit slope, and activation awareness is not applied through them.
Behind a max-pooling layer, the clean pooled slope is routed through each probe's own pooling winner, so that a probe which changes a window's maximum is credited to the unit it actually passed through.
\citet{singhal_how_2023} similarly exploit local activation derivatives, but use them to construct the gradient guess itself; here they only reweight the credit assigned by an otherwise unchanged perturbation estimator, and can therefore be layered on top of any of the methods in this work.
Activation awareness is an optional correction and is only used where explicitly indicated.

\begin{figure}[htbp]
    \centering
    \includegraphics[width=0.9\linewidth]{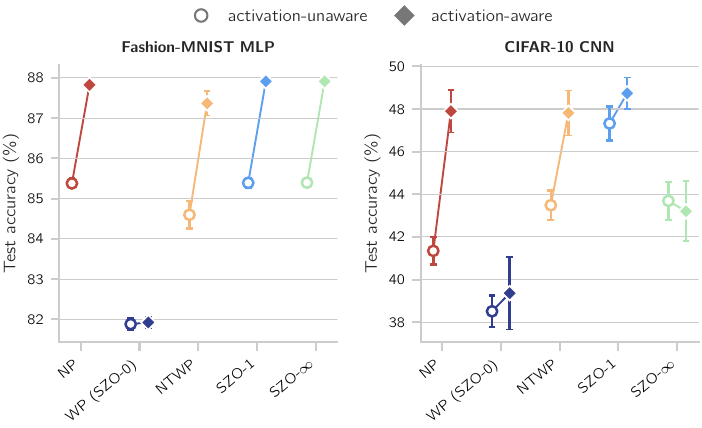}
    \caption{\textbf{Final test accuracy with and without activation awareness.}
    Open circles, activation-unaware; filled diamonds, activation-aware; SGD with \(n=1\), each setting at its own selected learning rate.
    Points show the mean over the last \(10\) of \(250\) epochs and \(10\) seeds, with \(95\%\) confidence intervals.}
    \label{fig:awareness}

    \vspace{1em}

    \centering
    \includegraphics[width=0.8\linewidth]{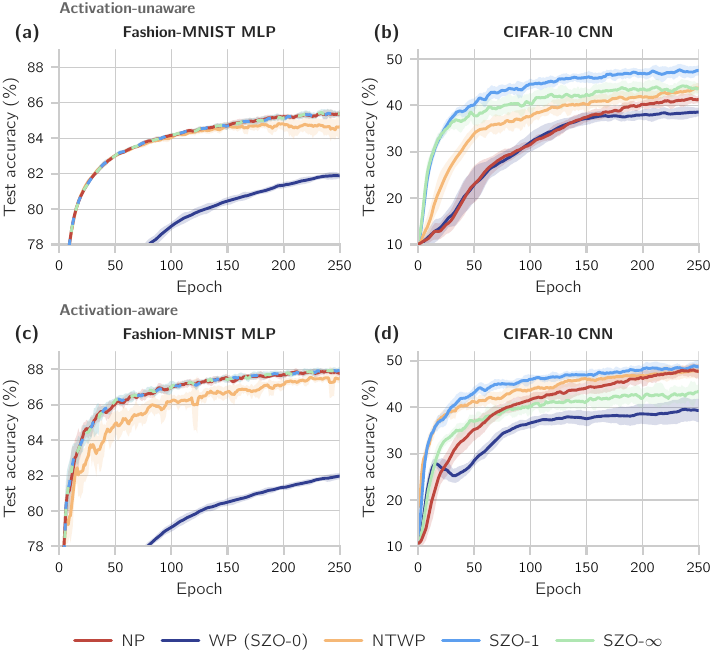}
    \caption{\textbf{Test accuracy over training without (a,b) and with (c,d) activation awareness.}
    Panels (a,b) repeat Figure~\ref{fig:performance}(a,c); BP (not shown) reaches \(89.9\%\) and \(83.6\%\).
    SGD with \(n=1\), each setting at its own selected learning rate; lines show the mean over \(10\) seeds, smoothed over \(5\) epochs, and shaded bands \(\pm 1\) standard deviation.}
    \label{fig:awareness-curves}
\end{figure}

\paragraph{Empirical effect.}
Figures~\ref{fig:awareness} and~\ref{fig:awareness-curves} compare every method with and without activation awareness, trained with SGD and \(n=1\), each setting at its own selected learning rate.
On the Fashion-MNIST MLP, awareness raises the final test accuracy of NP, NTWP, SZO-\(1\) and SZO-\(\infty\) and leaves WP unchanged.
On the CIFAR-10 CNN it helps NP  and NTWP most and SZO-\(1\) less, while WP and SZO-\(\infty\) stay within their confidence intervals.
All other experiments in this work are activation-unaware, except the bottom rows of Figures~\ref{fig:kladder} and~\ref{fig:lr-sweeps}.

\section{Bias-Variance Tradeoffs Across Perturbation Algorithms}
\label{app:bias-variance-tradeoffs}

All perturbation estimators trade unbiasedness against variance and coverage.
For a generic estimator, the linearised expected update can be written as
\[
  \mathbb{E}[\widehat{G}]
  =
  G M + O(\epsilon^2),
\]
where \(G\) is the exact gradient in the relevant matrix form and \(M\) is the effective covariance or spectral preconditioner induced by the perturbation rule.
Unstructured methods aim for \(M=I\), while SZO intentionally chooses \(M\) to emphasise data-supported directions.

\begin{table}[h]
\centering
\footnotesize
\caption{\textbf{Bias-variance structure of the perturbation estimators.} Here \(m\) is parameter dimension, \(k\) is activity dimension, \(q\) is layer-input dimension, \(C=AA^\top\), \(u_1\) is the leading eigenvector of \(C\), and \(n\) is the number of probes.}
\label{tab:bias-variance}
\begin{tabular}{@{}p{0.12\textwidth}p{0.30\textwidth}p{0.25\textwidth}p{0.24\textwidth}@{}}
\toprule
Method & Expected update geometry & Dominant variance scale & Main tradeoff \\
\midrule
WP/SPSA &
\(g\) when \(M_n\to I_m\) &
\(O(m/n)\|g\|^2\) &
Unbiased target, high dimension \\
NP &
\(\nabla_h\mathcal{L}\,J_{h,\theta}\) when \(M_\phi\to I_k\) &
\(O(k/n)\|\nabla_h\mathcal{L}\|^2\) &
Lower dimension, may be unrealisable \\
SZO-\(0\) &
\(G\) &
\(O(kq/n)\) nominal &
WP endpoint \\
SZO-\(1\) &
\(GC/\mathrm{Tr}(C)\) &
\(\propto \mathrm{Tr}(GC G^\top)/(n\,\mathrm{Tr}(C))\) &
Broad spectral bias, lower effective dimension \\
SZO-\(\infty\) &
\((Gu_1)u_1^\top\) &
\(O(k/n)\|Gu_1\|^2\) &
Lowest coverage, strongest variance reduction \\
NTWP &
\(GC\) up to normalisation (per-batch credit); \(G\) (per-sample) &
\(O(k_{\mathrm{eff}}/n)\) &
Realisable node-like credit \\
\bottomrule
\end{tabular}
\end{table}

For SZO-\(\kappa\), isotropic \(R\) gives
\[
  \mathbb{E}[\widehat{G}_{\kappa}]
  \propto
  G\frac{C^\kappa}{\mathrm{Tr}(C^\kappa)} .
\]
Thus the bias relative to the exact gradient is
\[
  \mathrm{Bias}_\kappa
  =
  G\left(\frac{C^\kappa}{\mathrm{Tr}(C^\kappa)}-I_q\right)
  +O(\epsilon^2).
\]
This bias is intentional: directions with little activation support receive less perturbation budget.
The variance decreases as the effective searched spectrum shrinks.
A convenient summary is the spectral effective dimension
\[
  q_{\mathrm{eff}}^{(\kappa)}
  =
  \frac{\mathrm{Tr}(C^\kappa)^2}{\mathrm{Tr}(C^{2\kappa})},
\]
which is large for flat spectra and small for concentrated spectra.
Increasing \(\kappa\) usually reduces variance by reducing \(q_{\mathrm{eff}}^{(\kappa)}\), but increases bias by discarding low-variance feature directions.
In closed form, with \(\tau=\mathrm{cond}(C)^\kappa\), the worst-case alignment obeys the Kantorovich bound \(\cos(G,GC^\kappa)\ge 2\sqrt{\tau}/(1+\tau)\), which loosens as \(\kappa\) grows (growing bias budget), while the relative variance scales as \(k\,q_{\mathrm{eff}}^{(\kappa)}/n\) with an exact \(1/n\) dependence (measured successive ratios \(1.96/2.00/2.00\) for \(n=1,2,4,8\)) and predictions tight to measurement within \(8\)--\(27\%\). We state bounds on estimator quality, not convergence guarantees.

NTWP has a related but implementation-friendly bias.
It perturbs parameters isotropically, but assigns credit through the induced activity perturbation:
\[
  \Delta H=RA,\qquad E=\Delta H A^\top=RC.
\]
Therefore, with per-batch credit, its expected update carries the same \(C\)-shaped geometry as SZO-\(1\), while its variance is governed by the effective rank of induced activity perturbations rather than the raw parameter dimension.
This is the central practical advantage: NTWP keeps perturbations realisable, respects shared parameters, and obtains node-like variance scaling when the activity locus is much smaller than the parameter space.

\section{Avoiding the Cost of Eigenvector Decompositions}
\label{iswp_details}

The SZO-\(\kappa\) family requires forming \(C^{\kappa/2}\) with \(C=AA^\top\), which in general demands an eigendecomposition of the \(q\times q\) feature covariance at every step (\(\mathcal{O}(q^3)\) cost).
For modern wide layers this can easily dominate the forward pass and undermines the principal motivation of zeroth-order methods, namely cheap, inference-only updates.
This appendix details the two regimes in which we are able to avoid the eigendecomposition entirely while still implementing the SZO-\(\infty\) (rank-1) estimator. It closes with the corresponding decomposition-free constructions for \(\kappa=1\) and \(\kappa=2\).
As a reminder, this SZO-\(\infty\) estimator makes use of a rank-1 perturbation \(\Delta W=\phi\,\hat{a}^\top\) constructed ideally from the leading eigenvector \(\hat{a}\) of the input features and a fresh isotropic channel noise \(\phi\in\mathbb{R}^{k}\) over the layer's outputs.

\paragraph{Fully connected layers: per-sample representatives.}
Recall that for a single-sample input (\(A=a\)) the leading eigenvector of \(C=aa^\top\) is exactly \(u_1=a/\|a\|_2\), so SZO-\(\infty\) collapses to NP.
For a mini-batch of \(B>1\) inputs \(\{a_b\}_{b=1}^B\), no eigendecomposition is needed if we are willing to treat each batch element as carrying its own rank-1 perturbation: we set
\[
  \hat{a}_b \;=\; a_b\big/\|a_b\|_2,\qquad
  \Delta W_b \;=\; \phi_b\,\hat{a}_b^\top,
\]
with independent channel noise \(\phi_b\in\mathbb{R}^{k_{\mathrm{out}}}\) drawn per (probe, batch element) pair.
The forward perturbation is computed without ever forming the dense weight perturbation, exploiting the rank-1 structure
\[
  \Delta h_{i,b} \;=\; \Delta W_b\,a_{i,b} \;=\; \phi_b\,\langle \hat{a}_b,\,a_{i,b}\rangle,
\]
where \(a_{i,b}\) is the noisy input replica for probe \(i\) and batch element \(b\).
The resulting estimator costs \(\mathcal{O}(B(k_{\mathrm{in}}+k_{\mathrm{out}}))\) per probe and never instantiates \(C\), let alone its decomposition.

\paragraph{Convolutional layers: an energy-weighted representative patch.}
Convolutions complicate the picture because weight sharing means that even a single image yields a high-rank \(C\): the unfolded patches from a single sample already span a many-dimensional subspace.
Here we cannot collapse SZO-\(\infty\) to NP as in the linear case, but we can still avoid the eigendecomposition by choosing \(\hat{a}_b\) to be a single representative \emph{patch} per sample.
Concretely, for each batch element \(b\) we collect the \(L=H_{\mathrm{out}}W_{\mathrm{out}}\) input patches \(\{x_{b,\ell}\in\mathbb{R}^{d}\}_{\ell=1}^{L}\) seen by the kernel (with \(d=C_{\mathrm{in}}\,k_H\,k_W\)), and from them form a single representative patch \(\hat{a}_b\in\mathbb{R}^{d}\) by normalising their energy-weighted mean,
\[
  \bar{a}_b \;=\; \frac{\sum_{\ell}\|x_{b,\ell}\|_2^{2}\,x_{b,\ell}}{\sum_{\ell}\|x_{b,\ell}\|_2^{2}},
  \qquad
  \hat{a}_b \;=\; \bar{a}_b\big/\|\bar{a}_b\|_2 .
\]
This is a cheap surrogate for the leading eigenvector of the per-sample patch second moment \(C_b=\sum_{\ell} x_{b,\ell}x_{b,\ell}^\top\).
One power-iteration step from the summed patches would weight each patch by its overlap with all patches, \(C_b\sum_m x_{b,m}=\sum_\ell\langle x_{b,\ell},\sum_m x_{b,m}\rangle\,x_{b,\ell}\); the energy-weighted mean keeps only each patch's overlap with itself, \(\|x_{b,\ell}\|_2^2\), and so favours high-energy patches at \(\mathcal{O}(dL)\) cost rather than the \(\mathcal{O}(d^3)\) of a full eigendecomposition.
A fresh isotropic channel noise \(\phi_b\in\mathbb{R}^{C_{\mathrm{out}}}\) is then drawn per (probe, batch element) pair.
The kernel perturbation is the rank-1 outer product \(\Delta K_b=\phi_b\,\hat{a}_b^\top\), reshaped to \((C_{\mathrm{out}},C_{\mathrm{in}},k_H,k_W)\).
A convolution applies its kernel to the patch at every spatial position \(\ell\), so the induced output perturbation at position \(\ell\) is
\[
  \Delta K_b\,x_{b,\ell} \;=\; \phi_b\,\langle \hat{a}_b,\,x_{b,\ell}\rangle ,
\]
and it can be evaluated without ever materialising \(\Delta K_b\): a single inner product per spatial position, broadcast across output channels by \(\phi_b\).
This costs \(\mathcal{O}(BLd + BC_{\mathrm{out}}L)\) per probe and, again, never instantiates either \(C\) or any decomposition of it.

\paragraph{SZO-\(1\) with per-sample credit: drawing the shaped probe directly.}
With per-sample credit, each batch element \(b\) receives its own probe shaped by its own second-moment matrix, \(\Psi_b=R_b\,C_b^{1/2}/\mathrm{Tr}(C_b)^{1/2}\).
Stacking the sample's \(L\) receptive-field patches as the rows of \(P_b\in\mathbb{R}^{L\times d}\) gives \(P_b^\top P_b=C_b\) and \(\|P_b\|_F^2=\mathrm{Tr}(C_b)\).
For Gaussian noise \(G_b\in\mathbb{R}^{k\times L}\) with independent \(\mathcal{N}(0,1/k)\) entries, each row of \(G_bP_b\) is therefore Gaussian with covariance \(P_b^\top P_b/k=C_b/k\), exactly as each row of \(R_bC_b^{1/2}\), so
\[
  \Psi_b \;\overset{d}{=}\; \frac{G_b\,P_b}{\|P_b\|_F}.
\]
The probe is thus drawn over the sample's patch positions and projected through its patches, at the cost of a matrix multiplication and with no eigendecomposition.
For a linear layer \(L=1\), and this reduces to the rank-1 per-sample construction above.

\paragraph{SZO-\(2\) requires no matrix root.}
At \(\kappa=2\) the shaping matrix is \(C^{\kappa/2}=C=AA^\top\) itself, so \(RC=(RA)A^\top\) is two matrix products, and the normaliser \(\mathrm{Tr}(C^2)=\|A^\top A\|_F^2\) is computed from the \(B\times B\) Gram matrix; neither \(C\) nor a decomposition of it needs to be formed.

\paragraph{Newton--Schulz approximates \(C^{1/2}\) without loss of alignment.}
Where an explicit \(C^{1/2}\) is needed (\(\kappa=1\)), three Newton--Schulz iterations (three matrix multiplies, no eigendecomposition) reach \(16\%\) relative matrix error yet cost only \(\sim\!1\%\) of gradient alignment (\(0.8917\) vs.\ \(0.9012\)); five iterations are indistinguishable from exact eigendecomposition. The estimator is insensitive to approximation error in the shaping matrix.

\section{Layer-wise Gradient Alignment}
\label{app:layerwise-alignment}

Figure~\ref{fig:angle_alignments} reports gradient alignment for a single layer of the networks that we investigated for training on the Fashion-MNIST and CIFAR-10 tasks.
This appendix extends that measurement to every layer, comparing the two credit regimes of Section~\ref{sec:credit-granularity}.
For \(20\) different random seeds, the networks are freshly initialised and a batch of \(B=128\) examples is drawn from the validation split.
For each layer in turn, only that layer is perturbed (all other layers are left clean), and its gradient estimate is formed from \(n\in\{1,2,4,\ldots\}\) probes (up to \(4096\) for the MLP and \(1024\) for the CNN) with perturbation scale \(\epsilon=10^{-4}\).
We report the angle between the estimated gradient and the exact backpropagation gradient for the same weights and batch.
Under \emph{per-batch} credit, each probe is shared by the whole batch and contributes a single scalar loss difference.
Under \emph{per-sample} credit, each batch element receives its own probe and its own loss difference.
Per-sample WP requires an independent weight perturbation per probe and per batch element, and is omitted because of its memory cost.

\subsection{Fashion-MNIST MLP}
\label{app:layerwise-fmnist}

\begin{figure}[htbp]
    \centering
    \includegraphics[width=1.0\linewidth]{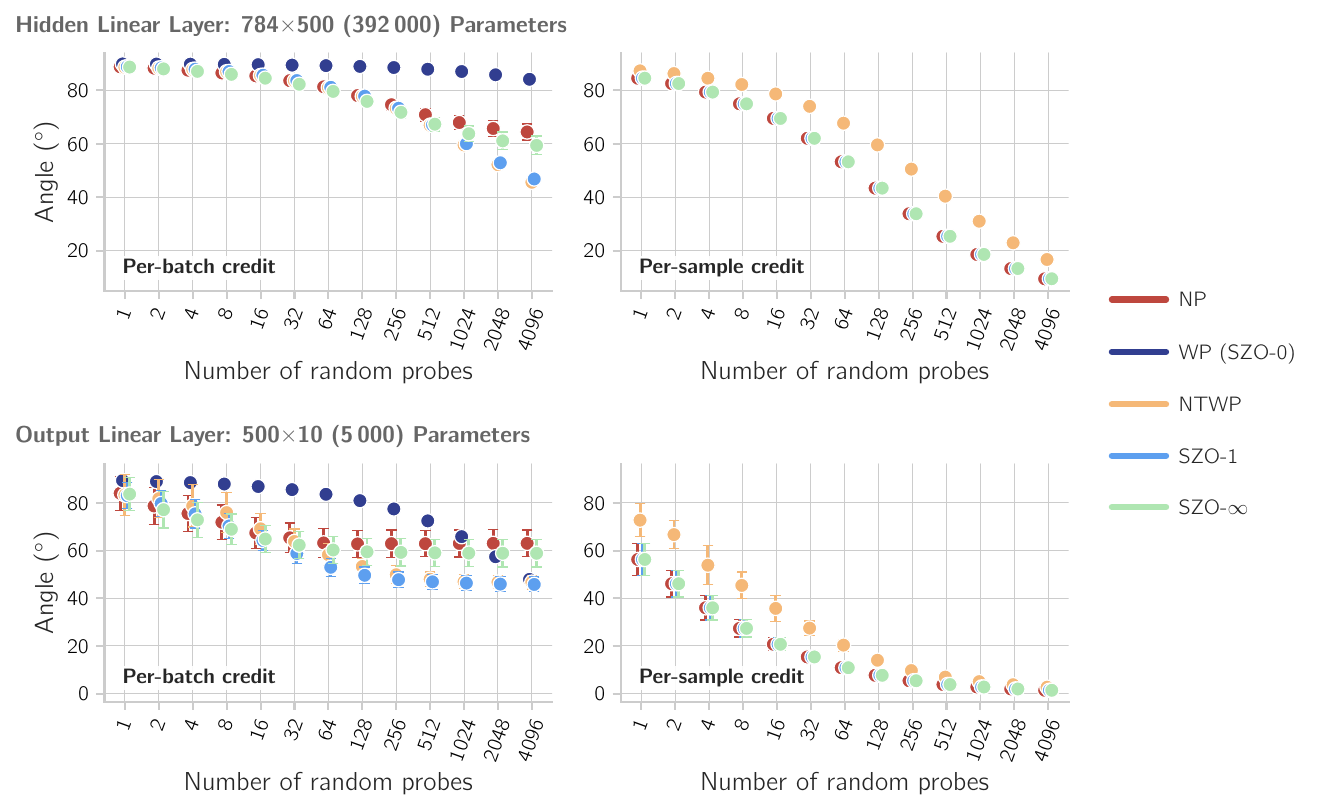}
    \caption{\textbf{Layer-wise gradient alignment for the Fashion-MNIST MLP at initialisation.}
    Each row is one weight layer; left panels use per-batch loss-difference credit and right panels per-sample credit.
    Points show the mean angle between the estimated and the backpropagation gradient over \(20\) seeds (initialisation and batch of \(128\)); error bars are \(\pm 1\) standard deviation.
    Only the layer shown is perturbed.
    On a linear layer per-sample SZO-\(\kappa\) is the same estimator for every \(\kappa>0\), so per-sample SZO-\(1\) and SZO-\(\infty\) show identical values; per-sample WP is omitted.}
    \label{fig:layerwise-fmnist}
\end{figure}

Figure~\ref{fig:layerwise-fmnist} shows both weight layers of the \(784{-}500{-}10\) MLP.
For a linear layer, each example's own second-moment matrix \(a_ba_b^\top\) is rank one, so every per-sample SZO-\(\kappa\) with \(\kappa>0\) reduces to the same estimator (Appendix~\ref{app:equivalences}), so per-sample SZO-\(1\) and SZO-\(\infty\) show identical values.
As predicted, per-sample SZO-\(1\), SZO-\(\infty\) and NP coincide at every \(n\) in both layers.
Under per-batch credit, NTWP and SZO-\(1\) are likewise indistinguishable, consistent with our theoretical analysis of their similarity (Appendix~\ref{app:equivalences}).

\subsection{CIFAR-10 CNN}
\label{app:layerwise-cifar}

\begin{figure}[htbp]
    \centering
    \includegraphics[width=0.75\linewidth]{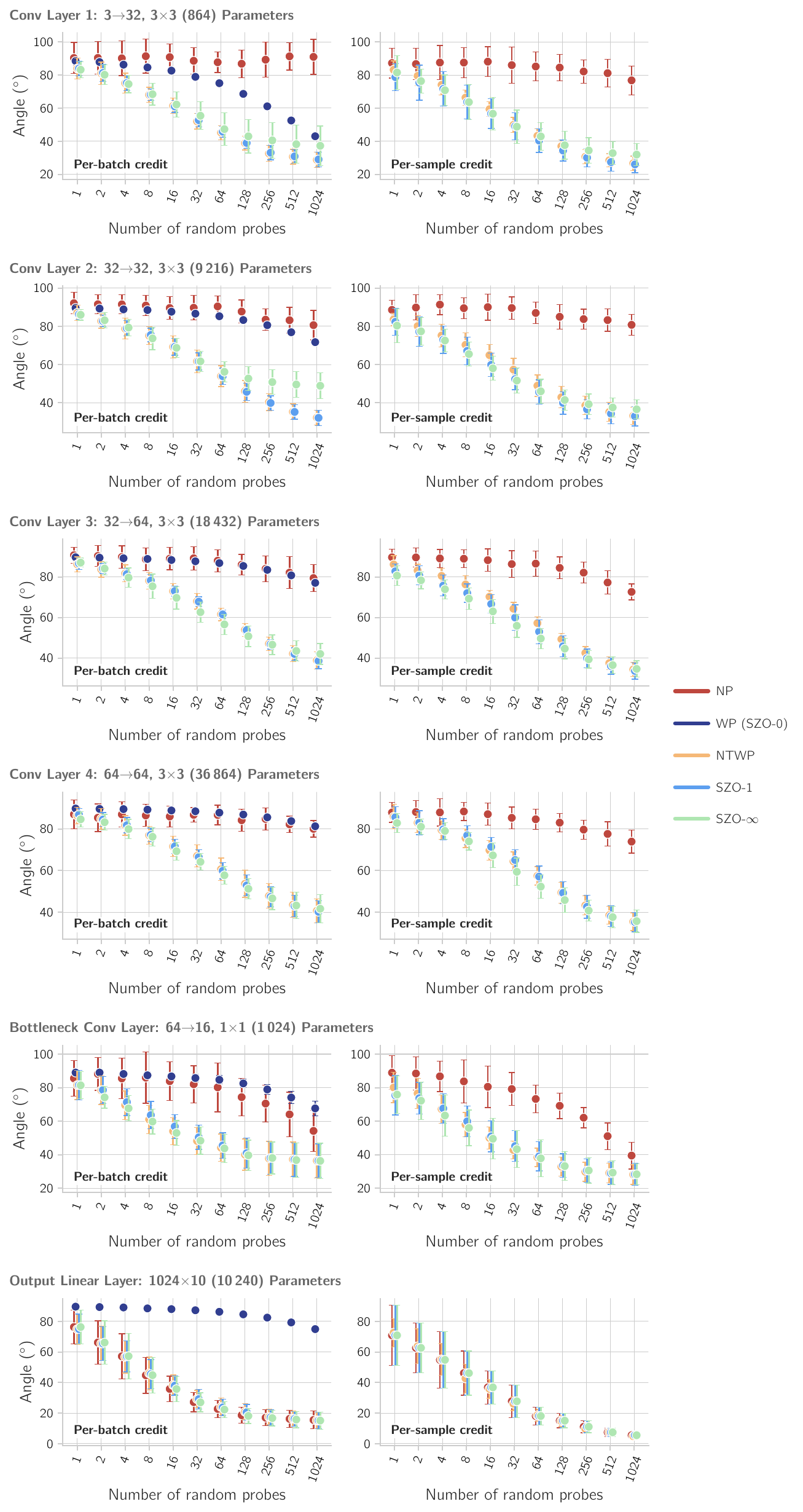}
    \caption{\textbf{Layer-wise gradient alignment for the CIFAR-10 CNN at initialisation.}
    Each row is one weight layer; left panels use per-batch loss-difference credit and right panels per-sample credit.
    Points show the mean angle between the estimated and the backpropagation gradient over \(20\) seeds (initialisation and batch of \(128\)); error bars are \(\pm 1\) standard deviation.
    Only the layer shown is perturbed.
    On the linear readout per-sample SZO-\(\kappa\) is the same estimator for every \(\kappa>0\), so per-sample SZO-\(1\) and SZO-\(\infty\) show identical values; per-sample WP is omitted.}
    \label{fig:layerwise-cifar}
\end{figure}

Figure~\ref{fig:layerwise-cifar} shows all six weight layers of the CIFAR-10 CNN: two \(3\times3\) convolutions with \(32\) channels, max-pooling, two \(3\times3\) convolutions with \(64\) channels, max-pooling, a \(1\times1\) bottleneck convolution to \(16\) channels, and a linear readout from the resulting \(1024\) features to the \(10\) classes.
On convolutional layers each example contributes many receptive-field patches, so its second-moment matrix is high rank and per-sample SZO-\(1\) and SZO-\(\infty\) are distinct estimators; on the linear readout they again coincide with per-sample NP.

NP aligns poorly on every \(3\times3\) convolutional layer under either credit, consistent with independent per-position node perturbations not being realisable by a shared kernel.
In per-batch credit NTWP and SZO-\(1\) are indistinguishable on every layer.

\section{Spectral Exponent Sweep}
\label{app:kappa-sweep}

The main text considers SZO-\(\kappa\) at \(\kappa\in\{0,1,\infty\}\).
To show how alignment depends on \(\kappa\) between and beyond these values, we sweep \(\kappa\in\{0,0.1,0.2,\ldots,0.9,1,2,5,10,\infty\}\) on the two layers of Figure~\ref{fig:angle_alignments}: the first layer of the Fashion-MNIST MLP and the last \(3\times3\) convolutional layer of the CIFAR-10 CNN.
The protocol is that of Appendix~\ref{app:layerwise-alignment} with the number of probes varied from \(n=1\) to \(n=256\): for each of \(20\) seeds the network is freshly initialised, a batch of \(B=128\) validation examples is drawn, only the measured layer is perturbed, and the angle to the exact backpropagation gradient is recorded under per-batch and per-sample credit.
\(\kappa=0\) is run as WP (\(C^0=I\)), \(\kappa=\infty\) as SZO-\(\infty\), and every other value as SZO-\(\kappa\) with the matrix power \(C^{\kappa/2}\).
All values of \(\kappa\) share the same initialisation, batch and random draws within a seed.

\begin{figure}[htbp]
    \centering
    \includegraphics[width=1.0\linewidth]{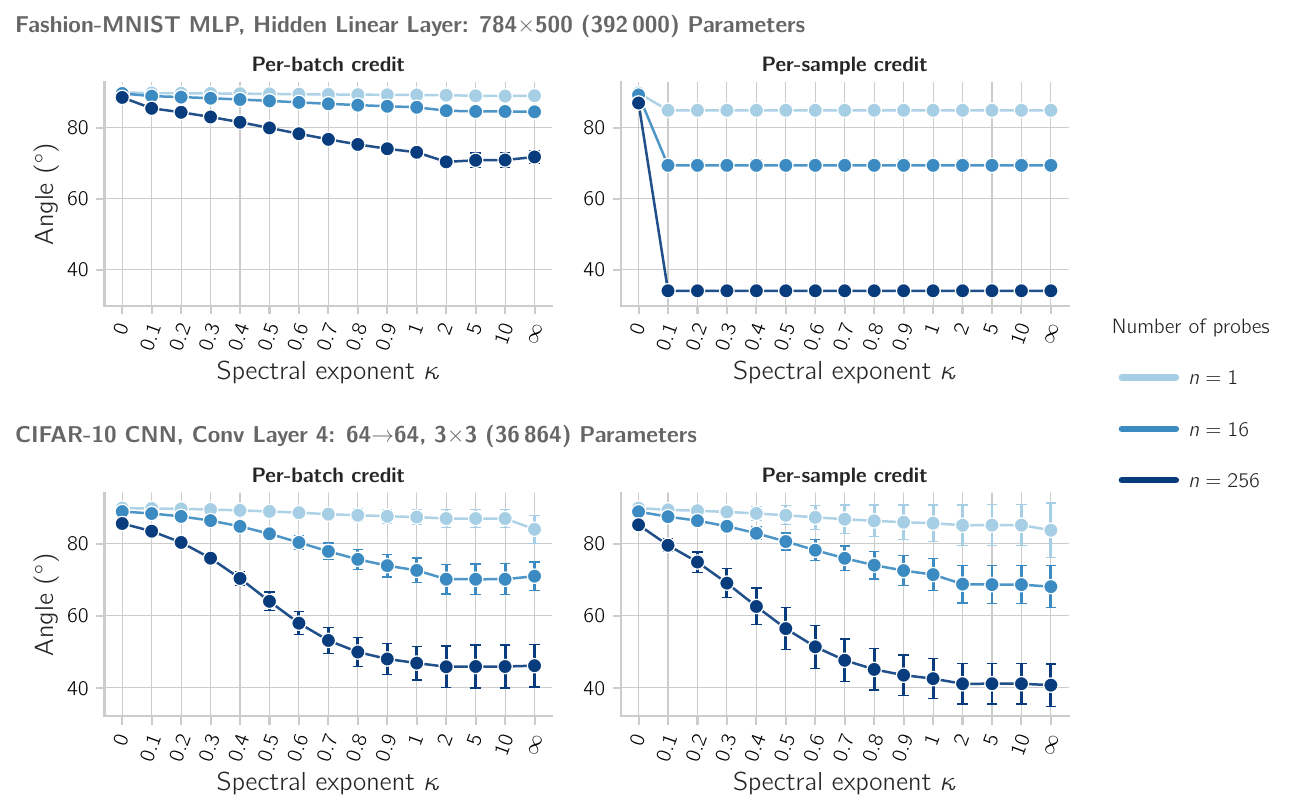}
    \caption{\textbf{Gradient alignment as a function of the spectral exponent \(\kappa\) at various probe budgets.}
    Rows show the first layer of the Fashion-MNIST MLP and the last \(3\times3\) convolutional layer of the CIFAR-10 CNN; left panels use per-batch loss-difference credit and right panels per-sample credit.
    Points show the mean angle between the estimated and the backpropagation gradient over \(20\) seeds (initialisation and batch of \(128\)); error bars are \(\pm 1\) standard deviation.
    \(\kappa=0\) (WP), \(\kappa=1\) and \(\kappa=\infty\) are highlighted in their method colours; the \(\kappa\) axis is categorical.}
    \label{fig:kappa-sweep}
\end{figure}

Figure~\ref{fig:kappa-sweep} shows the result.
On the linear layer under per-sample credit, the angle falls from approximately \(90^\circ\) at \(\kappa=0\) to a significantly lower value but then remains identical for every \(\kappa>0\), as each example's second-moment matrix is rank one and SZO-\(\kappa\) reduces to the same estimator (Appendix~\ref{app:equivalences}).
Under per-batch credit on the same layer, \(\kappa\) has a small effect at this budget: the angle decreases smoothly from approximately \(90^\circ\) but in some cases appears to rise for large \(\kappa\).
Convolutional layers show a very similar effect, though now (as the theory predicts) there are significant differences between the alignment for different spectral exponents \(\kappa > 0\) even in the per-sample case.
On both layers, alignment improves steadily up to \(\kappa=1\) and saturates by \(\kappa\approx2\), with no intermediate value outperforming the plateau; \(\kappa\in\{0,1,\infty\}\) therefore span the behaviour of the family.
On the convolutional layer, SZO-\(\infty\) lies within one standard deviation of the large-\(\kappa\) plateau; it is implemented with the energy-weighted patch surrogate of Appendix~\ref{iswp_details} rather than the exact \(\kappa\to\infty\) limit.

\begin{figure}[htbp]
    \centering
    \includegraphics[width=1.0\linewidth]{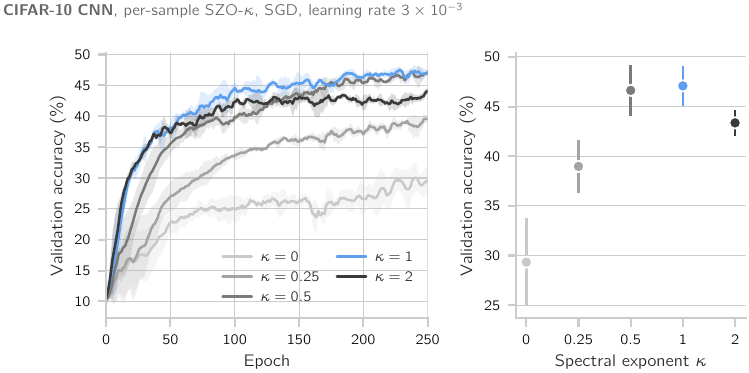}
    \caption{\textbf{Training the CIFAR-10 CNN with per-sample SZO-\(\kappa\) for different spectral exponents \(\kappa\).}
    Left, validation accuracy over training (mean over \(3\) seeds, smoothed over \(5\) epochs, shaded bands \(\pm 1\) standard deviation); right, its mean over the last \(10\) of \(250\) epochs with \(95\%\) confidence intervals.
    SGD with \(n=1\) at the learning rate of SZO-\(1\) (\(3\times10^{-3}\)) for every \(\kappa\); \(\kappa=1\) is SZO-\(1\) and shown in its method colour, and \(\kappa=0\) denotes the limit \(\kappa\to0^{+}\) (see text).}
    \label{fig:kappa-training}
\end{figure}

\paragraph{Training.}
To test whether the alignment trend carries over to training, we train the CIFAR-10 CNN with per-sample SZO-\(\kappa\) for \(\kappa\in\{0,0.25,0.5,1,2\}\), using SGD with \(n=1\) and the learning rate selected for SZO-\(1\) (\(3\times10^{-3}\)) for every \(\kappa\), with \(3\) seeds.
Figure~\ref{fig:kappa-training} shows that validation accuracy after \(250\) epochs rises with \(\kappa\) up to \(\kappa=1\) (\(46.6\%\) at \(\kappa=0.5\), \(47.1\%\) at \(\kappa=1\)) and falls again at \(\kappa=2\) (\(43.4\%\)), so unlike alignment at initialisation, final accuracy does not saturate beyond \(\kappa=1\).
Every \(\kappa\) uses the learning rate tuned for \(\kappa=1\), which may favour values near \(\kappa=1\) and partly explain the drop at \(\kappa=2\).
For \(\kappa=0\) the implementation takes the limit \(\kappa\to0^{+}\) (Appendix~\ref{app:equivalences}), in which \(C_b^{\kappa/2}\) becomes the projector onto the span of the sample's own patches: this is per-sample WP only on layers whose patches span the layer input, noise restricted to the patch span on the other convolutions, and NP on the linear readout, so it is not the per-batch WP of the other experiments.

\section{Final Test Accuracies}
\label{app:final-accuracies}

\begin{figure}[htbp]
    \centering
    \includegraphics[width=0.9\linewidth]{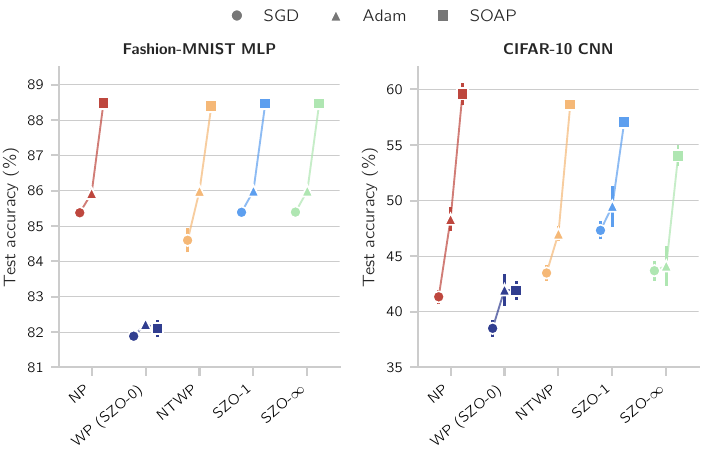}
    \caption{\textbf{Final test accuracy per method and optimiser.}
    SGD (circles, \(10\) seeds), Adam (triangles) and SOAP (squares, \(5\) seeds each), \(n=1\), each method at its own selected learning rate.
    Points show the mean over the last \(10\) of \(250\) epochs with \(95\%\) confidence intervals.
    Learning rates were selected on a validation split (20\% of the training set; Table~\ref{tab:lrs}); the final runs train on the full training set and are evaluated on the official test split.
    BP (not shown) reaches \(89.9\), \(90.2\) and \(90.4\%\) on Fashion-MNIST and \(83.6\), \(84.2\) and \(83.7\%\) on CIFAR-10 with SGD, Adam and SOAP.}
    \label{fig:optimizers}

    \vspace{1em}

    \centering
    \includegraphics[width=0.8\linewidth]{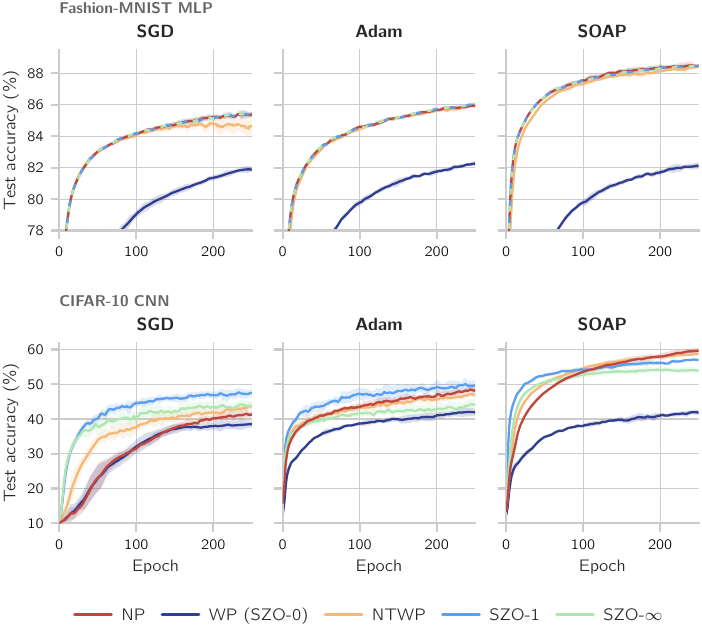}
    \caption{\textbf{Test accuracy over training for each optimiser.}
    Rows show the Fashion-MNIST MLP and the CIFAR-10 CNN, columns SGD, Adam and SOAP, with each method at its own selected learning rate and \(n=1\).
    Lines show the mean over seeds (SGD \(10\), Adam and SOAP \(5\)), smoothed over \(5\) epochs, and shaded bands \(\pm 1\) standard deviation; BP is not shown (final accuracies in Figure~\ref{fig:optimizers}).}
    \label{fig:curves-optimizers}
\end{figure}

Figure~\ref{fig:optimizers} reports the final test accuracies of the perturbation-based learning methods on Fashion-MNIST and CIFAR-10, trained with SGD as in Figure~\ref{fig:performance} and additionally with Adam and SOAP, each at its own selected learning rate (Table~\ref{tab:lrs}).
As expected, BP achieves the highest accuracy on both datasets (\(89.9\), \(90.2\) and \(90.4\%\) on Fashion-MNIST and \(83.6\), \(84.2\) and \(83.7\%\) on CIFAR-10 with SGD, Adam and SOAP).
Among the perturbation-based methods, NP, SZO-\(1\) and SZO-\(\infty\) perform identically on Fashion-MNIST with NTWP close behind, while SZO-\(1\) obtains the highest accuracy on CIFAR-10 with SGD and Adam.
With SOAP, every method except WP improves substantially, most of all on CIFAR-10, where NP and NTWP reach the highest accuracies.
In contrast, WP performs substantially worse, especially on CIFAR-10, and gains little from adaptive optimisers.
Figure~\ref{fig:curves-optimizers} shows the corresponding training curves for each optimiser.


\section{Number of Probes per Step}
\label{app:probes}

\begin{figure}[htbp]
    \centering
    \includegraphics[width=1.0\linewidth]{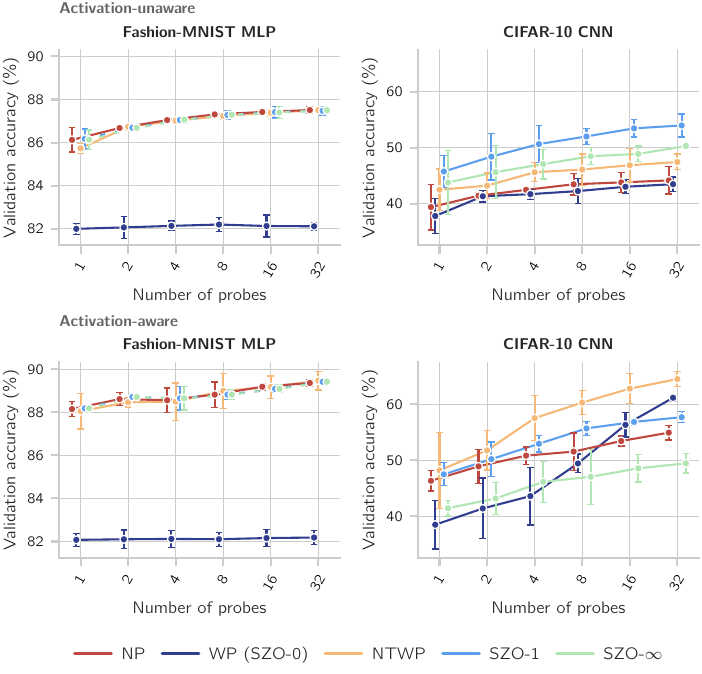}
    \caption{\textbf{Validation accuracy against the number of probes per step \(n\).}
    Top, after \(250\) epochs, as in Figure~\ref{fig:performance}(b,d); bottom, the same with activation awareness.
    SGD at each method's learning rate for \(n=1\) in that setting; points show the mean over \(3\) seeds with \(95\%\) confidence intervals.}
    \label{fig:kladder}
\end{figure}

\begin{figure}[htbp]
    \centering
    \includegraphics[width=1.0\linewidth]{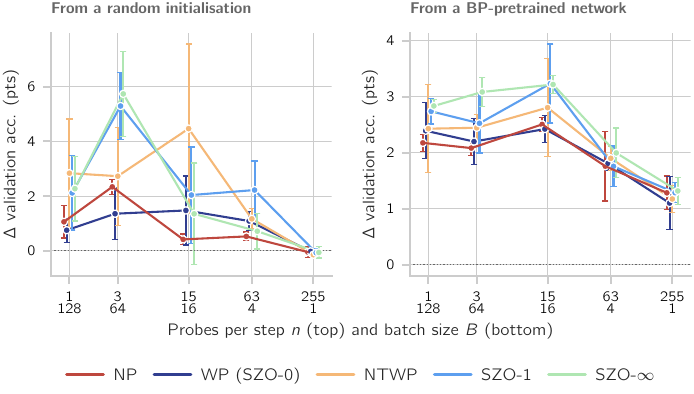}
    \caption{\textbf{Probes or examples at a fixed compute per step, measured.}
    Change of validation accuracy (points) of the CIFAR-10 CNN after \(1250\) SGD steps with \(n\) probes on a batch of \(B\) examples, where every step costs the same \(B(n+1)=256\) forward passes: the \(n\) perturbed passes and the clean pass of each example.
    Every run starts from the same weights: a random initialisation (left), or that initialisation after \(3\) epochs of BP (right, \(40.9\%\) validation accuracy).
    Each method uses the learning rate selected per \((n,B)\) from the pretrained start (lowest final validation loss, \(3\) seeds); points show the mean \(\pm\) one standard deviation over \(3\) further seeds.
    For reference, BP (\(B=128\)) changes the validation accuracy over the same \(1250\) steps by \(+35.8\pm0.8\) (left) and \(+9.8\pm1.2\) points (right).}
    \label{fig:bvp}

    \vspace{0.5em}
    \includegraphics[width=1.0\linewidth]{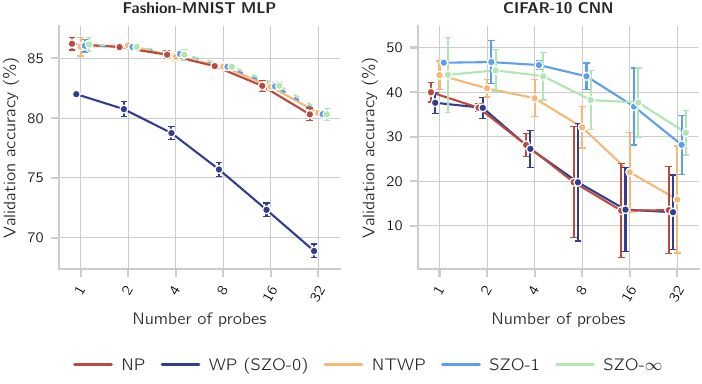}
    \caption{\textbf{Validation accuracy at an equal budget of forward passes.}
    The activation-unaware runs of Figure~\ref{fig:kladder} (top) after \(500/(n+1)\) epochs, so that every point has used the same number of forward passes, the \(n\) perturbed passes and the clean pass of each example, as \(n=1\) after \(250\) epochs.
    Points show the accuracy of that epoch, mean over \(3\) seeds with \(95\%\) confidence intervals.}
    \label{fig:equal-passes}
\end{figure}

Figure~\ref{fig:kladder} extends Figure~\ref{fig:performance}(b,d).
We compare three ways of paying for more probes: with more compute per step at a fixed batch (Figure~\ref{fig:kladder}), with fewer steps at an equal total of forward passes (Figure~\ref{fig:equal-passes}), and with fewer examples at a fixed compute per step (Figure~\ref{fig:bvp}).
Each method is trained with SGD at the learning rate selected for \(n=1\) (Table~\ref{tab:lrs}) with \(n\in\{1,2,4,8,16,32\}\) probes per step for \(250\) epochs on the validation split, with \(3\) seeds and the batch size fixed at \(128\).
The top row repeats the accuracy after \(250\) epochs; the bottom row repeats it with activation awareness (Appendix~\ref{app:activation-awareness}), each method at the learning rate selected for \(n=1\) in that setting (Figure~\ref{fig:lr-sweeps}).
With awareness, the methods whose perturbation is shared across the batch gain most on CIFAR-10 (\(n=1\) to \(32\): NTWP \(48.2\) to \(64.5\%\), WP \(38.5\) to \(61.2\%\)); awareness also raises their selected learning rates, which may account for part of this.
At an equal budget of forward passes, i.e.\ after \(500/(n+1)\) epochs, so that every point has used the same number of forward passes including the clean pass (Figure~\ref{fig:equal-passes}), more probes per step do not pay off: no method improves on \(n=1\) by more than a point, and every method falls for \(n\ge8\), on CIFAR-10 least the SZO-\(\kappa\) estimators (activation-unaware SZO-\(1\): \(46.6\%\) at \(n=1\), \(43.5\%\) at \(n=8\), \(28.1\%\) at \(n=32\)).

\paragraph{Probes or examples at a fixed compute per step.}
A fixed compute per step can buy either more probes or more examples.
In Figure~\ref{fig:bvp} every step costs the same \(B(n+1)=256\) forward passes, from \(n=1\), \(B=128\) to \(n=255\), \(B=1\), and every method takes the same \(1250\) SGD steps (\(4\) epochs at \(B=128\)), starting from a random initialisation and from a network pretrained by BP for \(3\) epochs.
From the pretrained network the trade is close to neutral down to \(B=16\), and only very small batches lose.
From the random initialisation every method is still close to chance (at most \(16\%\) validation accuracy), but NTWP and the SZO estimators already gain up to \(4.5\) to \(5.7\) points against at most \(2.3\) for NP and WP, in line with their alignment on convolutional layers (Figure~\ref{fig:angle_alignments}).
Over full training at the same compute per step as \(n=1\), \(B=128\), the trade pays off: SZO-\(1\) reaches \(48.7\%\) (\(n=4\), \(B=50\)) and \(49.9\%\) (\(n=16\), \(B=14\)) validation accuracy against \(45.7\%\), and WP \(40.5\%\) and \(41.2\%\) against \(37.8\%\) (mean over \(3\) seeds of the last \(4\%\) of training steps), with \(n=4\) and \(n=16\) within seed noise.
At a fixed learning rate the expected step is the same for every \(n\) and \(B\): the number of steps sets how far training gets and the noise of a step how close it settles, which is why trading steps for probes loses (Figure~\ref{fig:equal-passes}) while trading examples for probes gains.
In practice, this suggests choosing the estimator first, keeping the number of steps, and spending a fixed budget per step on a few probes on a smaller batch; on the CNN, \(n=4\) already gives most of the gain of \(n=16\).

\section{Iterative vs.\ Simultaneous Layer Perturbation}
\label{app:seq-vs-sim}

\begin{figure}[htbp]
    \centering
    \includegraphics[width=1.0\linewidth]{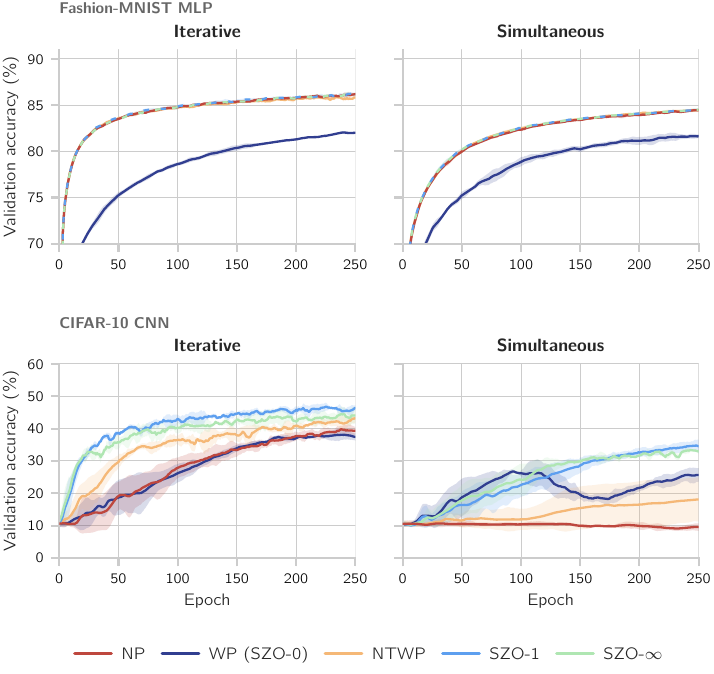}
    \caption{\textbf{Validation accuracy dynamics under iterative (one layer at a time) versus simultaneous (all layers perturbed at once) probing modes.}
    The learning rate of each method was selected separately for each mode by the same grid search; all other hyperparameters are matched between the two configurations.
    SGD with \(n=1\); lines show the mean over \(3\) seeds, smoothed over \(5\) epochs, and shaded bands \(\pm 1\) standard deviation.}
    \label{fig:seq-vs-sim}
\end{figure}

The estimators considered in this work can be applied in two different probing modes when training multi-layer networks.
In the \emph{iterative} mode, only one layer is perturbed at a time and the loss difference is used to update that layer alone, cycling through all layers within each training step.
In the \emph{simultaneous} mode, every layer is perturbed at the same time and a single loss difference is used to update all layers jointly.
Simultaneous probing is appealing because it requires only a single perturbed forward pass per step regardless of network depth, but it also raises the concern that noise injected at upstream layers will corrupt the activity at downstream layers and bias their estimates.
Figure \ref{fig:seq-vs-sim} compares the resulting validation accuracy dynamics under these two modes for the methods studied in this work.

As can be seen, on the Fashion-MNIST MLP the impact of iterative vs simultaneous training is qualitatively minimal, though the accuracies reached with iterative probing are up to two points higher.
On the CIFAR-10 CNN, simultaneous probing costs considerably more: SZO-\(1\) and SZO-\(\infty\) still reach \(35\%\) and \(33\%\) validation accuracy after \(250\) epochs, compared with \(46\%\) and \(44\%\) when iterative, while WP and NTWP learn slowly and NP does not improve on chance.
Both modes are compared at an equal number of training steps, at which simultaneous probing uses about half the compute of iterative probing on the CNN and three quarters on the MLP (Appendix~\ref{app:compute_hyperparams}, Computational complexity).
Therefore the iterative method was used across all simulations in this work.

\section{Computational Resources and Hyperparameters}
\label{app:compute_hyperparams}

Training and gradient-alignment runs used GPU nodes equipped with four accelerators each, AMD EPYC host CPUs, and high-bandwidth NVLink interconnect between the four GPUs on a node, as well as GPU workstations.
Each individual training run for the Fashion-MNIST MLP and the CIFAR-10 CNN occupied a single GPU.
The four GPUs of a node were used to run independent seeds and method configurations in parallel rather than to shard a single model.
Table~\ref{tab:compute} lists the compute, wall-clock time and peak memory per training step of every method, measured on a single NVIDIA A100.

\begin{table}[h]
\centering
\footnotesize
\caption{\textbf{Compute and memory per training step.} Batch size $128$, $n=1$, FP32, iterative probing, final configuration; MACs: multiply-accumulates per training step counted on our implementation, relative to BP on the same task (Computational complexity, below); time: measured time per step relative to BP, on one dedicated NVIDIA A100 (40\,GB), all methods in one session, median over $3$ repeated runs of the mean over epochs 2--3 (repeats differ by at most $3\%$), without activation awareness, which adds $3$ to $13\%$ (Appendix~\ref{app:activation-awareness}); peak: peak allocated GPU memory. MAC counts leave out element-wise operations, memory traffic and kernel-launch overhead, which dominate the step time of networks this small, so the measured time ratios are larger. Our implementations are research code and not optimised for speed; faster implementations of every method are likely possible.}
\label{tab:compute}
\begin{tabular}{lcccccccc}
\toprule
 & \multicolumn{4}{c}{Fashion-MNIST} & \multicolumn{4}{c}{CIFAR-10} \\
\cmidrule(lr){2-5}\cmidrule(lr){6-9}
\textbf{Method} & MACs & ms/step & time & peak MiB & MACs & ms/step & time & peak MiB \\
\midrule
BP & 1.0$\times$ & 2.1 & 1.0$\times$ & 77 & 1.0$\times$ & 7.1 & 1.0$\times$ & 573 \\
\midrule
NP & 2.0$\times$ & 6.2 & 2.9$\times$ & 69 & 2.0$\times$ & 19.2 & 2.7$\times$ & 563 \\
WP (SZO-0) & 2.0$\times$ & 6.7 & 3.1$\times$ & 70 & 1.7$\times$ & 19.8 & 2.8$\times$ & 343 \\
NTWP & 2.5$\times$ & 6.7 & 3.1$\times$ & 69 & 2.0$\times$ & 20.3 & 2.9$\times$ & 563 \\
SZO-1 & 2.0$\times$ & 6.8 & 3.2$\times$ & 69 & 2.5$\times$ & 27.7 & 3.9$\times$ & 388 \\
SZO-$\infty$ & 2.0$\times$ & 6.7 & 3.1$\times$ & 69 & 1.3$\times$ & 23.9 & 3.4$\times$ & 528 \\
\bottomrule
\end{tabular}

\end{table}

\paragraph{Computational complexity.}
We count multiply-accumulates (MACs) per training step.
A layer \(l\) with fan-in \(q_l\) (\(c_{\mathrm{in}}\) times the kernel area for a convolution), \(k_l\) outputs and \(P_l\) output positions (\(P_l=1\) for a linear layer) costs \(F_l=BP_lq_lk_l\) MACs for a forward pass over a batch of \(B\), and the network \(F=\sum_lF_l\); backpropagation costs about \(3F\).
A perturbation step costs a clean forward pass, the \(n\) probes, and a local update:
\[
  F
  \;+\;
  \begin{cases}
    n\bar{L}F & \text{iterative}\\
    nF & \text{simultaneous}
  \end{cases}
  \;+\;
  \text{update},
  \qquad
  \bar{L}=\frac{1}{F}\sum_{m=1}^{L}m\,F_m\in[1,L].
\]
An iterative probe of layer \(l\) reuses the cached clean input of that layer and reruns only the layers from \(l\) onwards, so \(\bar{L}\) is the depth weighted by where the compute lies: \(1.01\) for our MLP, whose compute sits almost entirely in the first layer, and \(2.93\) for our CNN.
The update is local to the perturbed layer and the same in both modes:

\begin{center}
\footnotesize
\begin{tabular}{@{}ll@{}}
\toprule
Method & Update cost \\
\midrule
NP, NTWP & \(\approx F\) (one weight-gradient product, \(\sum_b\Delta\mathcal{L}_b\,\phi_ba_b^\top\)) \\
WP & \(O(nq_lk_l)\) per layer \\
SZO-\(1\) (per-sample) & \(\approx nF\) (forming each example's probe from its patches) \\
SZO-\(\infty\) (per-sample) & \(O(BP_lq_l)\) for \(\hat{a}_b\), \(O(Bq_lk_l)\) for the update \\
\bottomrule
\end{tabular}
\end{center}

These expressions give the minimal work of each mode.
Table~\ref{tab:compute} instead counts the MACs of our implementation, together with measured time and peak memory.
Iterative probing runs one pass per perturbed layer, back to back, rather than all layers in one go, and each pass carries its own overhead: it recomputes the clean rows of the perturbed layer (\(F\) per step in total), and on linear layers WP and NTWP apply the weight perturbation as a separate product (\(nF_l\)).
Simultaneous probing pays this overhead in a single pass.
By the expressions above the two modes therefore cost the same on the MLP (\(\bar{L}\approx1\)), whereas in the counts of our implementation simultaneous probing uses about three quarters of the MACs of iterative probing on the MLP and about half on the CNN (\(0.4\) to \(0.5\times\); SZO-\(1\): \(0.86\times\)), at an equal number of steps.
In measured time the gap is smaller still: a simultaneous step takes \(0.9\) of an iterative one on the MLP and \(0.76\) to \(0.96\) on the CNN.
No method stores a backward graph, so peak memory is close to that of backpropagation; per-sample SZO-\(1\) additionally holds the \(nB\) per-example weight perturbations of one layer (\(nBq_lk_l\) entries).

We estimate the total compute used for the reported results, including hyperparameter exploration and discarded preliminary runs, at the order of a few hundred GPU-hours.

\subsection{Network Architectures}
\label{app:architectures}

Table~\ref{tab:architectures} lists the weight layers of both networks.
Every layer has a bias, and every layer except the output is followed by a leaky ReLU (negative slope \(0.01\)); no normalisation layers are used.
Convolutions use stride \(1\), and \(3\times3\) convolutions use padding \(1\).
In the CNN, \(2\times2\) max-pooling follows the second and fourth convolutions, and the \(1\times1\) bottleneck reduces the \(64\times8\times8\) feature map to \(16\times8\times8\), which is flattened into the \(1024\) inputs of the linear readout.

\begin{table}[h]
\centering
\footnotesize
\caption{\textbf{Network architectures.} Output shapes are channels \(\times\) height \(\times\) width; weight counts exclude biases.}
\label{tab:architectures}
\begin{tabular}{llllr}
\toprule
\textbf{Layer} & \textbf{Type} & \textbf{Configuration} & \textbf{Output} & \textbf{Weights} \\
\midrule
\multicolumn{5}{l}{\textit{Fashion-MNIST MLP (input \(784\))}} \\
1 & Linear & \(784\to500\) & \(500\) & \(392{,}000\) \\
2 & Linear (output) & \(500\to10\) & \(10\) & \(5{,}000\) \\
\midrule
\multicolumn{5}{l}{\textit{CIFAR-10 CNN (input \(3\times32\times32\))}} \\
1 & Conv \(3\times3\) & \(3\to32\) & \(32\times32\times32\) & \(864\) \\
2 & Conv \(3\times3\) + max-pool & \(32\to32\) & \(32\times16\times16\) & \(9{,}216\) \\
3 & Conv \(3\times3\) & \(32\to64\) & \(64\times16\times16\) & \(18{,}432\) \\
4 & Conv \(3\times3\) + max-pool & \(64\to64\) & \(64\times8\times8\) & \(36{,}864\) \\
5 & Conv \(1\times1\) (bottleneck) & \(64\to16\) & \(16\times8\times8\) & \(1{,}024\) \\
6 & Linear (output) & \(1024\to10\) & \(10\) & \(10{,}240\) \\
\bottomrule
\end{tabular}
\end{table}

\subsection{Hyperparameters}

\begin{figure}[htbp]
    \centering
    \includegraphics[width=1.0\linewidth]{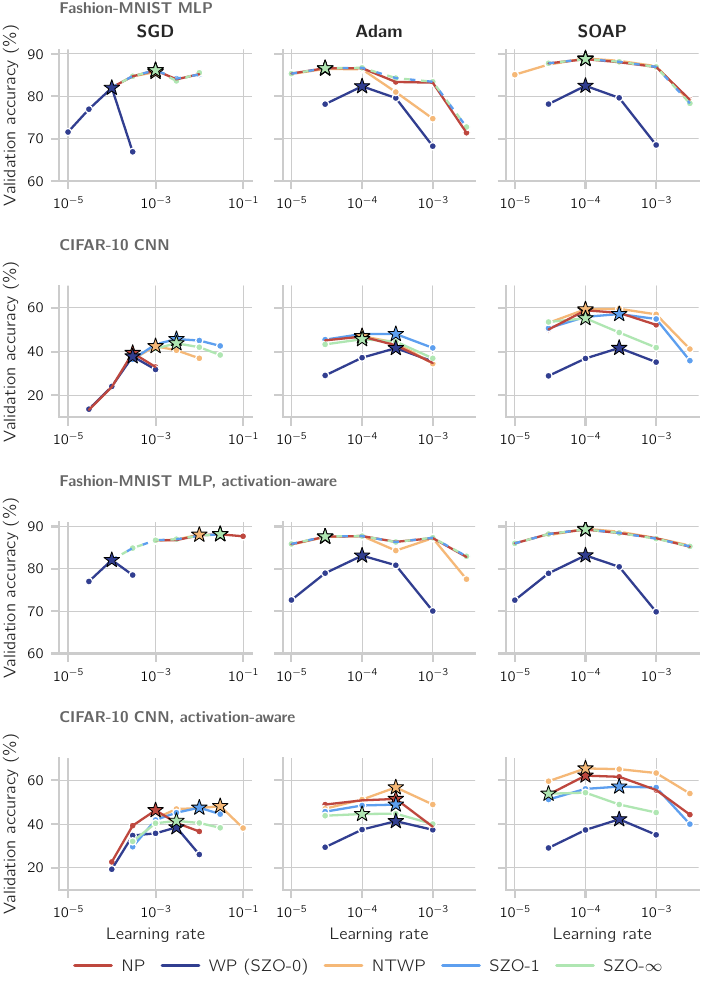}
    \caption{\textbf{Validation accuracy across the learning-rate grid}, activation-unaware (top two rows) and activation-aware (bottom two rows, Appendix~\ref{app:activation-awareness}).
    Mean over the last \(10\) of \(250\) epochs and \(3\) seeds, \(n=1\), for SGD, Adam and SOAP; stars mark the selected learning rate (Table~\ref{tab:lrs}).
    Learning rates at which at least one seed diverged are omitted.}
    \label{fig:lr-sweeps}
\end{figure}

Table~\ref{tab:hparams} summarises the hyperparameters used for the Fashion-MNIST MLP and CIFAR-10 CNN experiments.
The learning rate was selected by a grid search for each method, optimiser and dataset (Table~\ref{tab:lrs}, Figure~\ref{fig:lr-sweeps}); all other hyperparameters are shared.
Optimiser-specific settings for Adam and SOAP, as well as per-method probing settings such as the number of probes, perturbation scale $\epsilon$, SZO-$\kappa$ value, and simultaneous versus iterative probing, are listed where they differ from the defaults of the corresponding optimiser or estimator.
 
\begin{table}[h]
\centering
\footnotesize
\caption{\textbf{Overview of hyperparameters used for the Fashion-MNIST MLP and CIFAR-10 CNN experiments.} Shared settings are listed once; dataset- or architecture-specific settings are indicated where they differ.}
\label{tab:hparams}
\begin{tabular}{@{}ll>{\raggedright\arraybackslash}p{0.56\linewidth}@{}}
\toprule
\textbf{Hyperparameter} & \textbf{Symbol} & \textbf{Value} \\
\midrule
\multicolumn{3}{l}{\textit{Architecture and data}} \\
\midrule
Architecture Fashion-MNIST &  & MLP, $784{-}500{-}10$ \\
Architecture CIFAR-10 &  & CNN, $[32,32,\mathrm{pool},64,64,\mathrm{pool},16_{1\times1}]\to 10$ \\
Activation function &  & Leaky ReLU \\
Preprocessing &  & Fashion-MNIST: normalise; CIFAR-10: normalise, random crop (padding 4), horizontal flip \\
Loss &  & Cross-entropy with label smoothing $0.1$ \\
Data splits &  & Calibration: 80/20 split of the training set; final runs: full training set, test set \\
\midrule
\multicolumn{3}{l}{\textit{Training setup}} \\
\midrule
Batch size & $B$ & $128$ \\
Epochs &  & $250$ \\
Random seeds &  & $3$ per learning rate (calibration); $10$ (SGD) and $5$ (Adam, SOAP) final runs; $20$ gradient alignment \\
Numerical precision &  & FP32 (TF32 off) \\
Gradient clipping &  & Adaptive, per parameter tensor: norm capped at $5\times$ the (lower) median of the previous $100$ norms (after $50$ warm-up steps); perturbation methods only \\
\midrule
\multicolumn{3}{l}{\textit{Optimisation}} \\
\midrule
Optimiser &  & SGD without momentum; Adam and SOAP in Appendix~\ref{app:final-accuracies} \\
Learning rate & $\alpha$ & Per method, optimiser and task (Table~\ref{tab:lrs}); constant schedule \\
Learning-rate selection &  & Half-decade grid; lowest validation loss over the last $10$ epochs among learning rates where every seed completed \\
Weight decay & $\lambda$ & $0$ \\
\midrule
\multicolumn{3}{l}{\textit{Perturbation and probing}} \\
\midrule
Number of probes per step & $n$ & $1$, unless specified otherwise \\
Perturbation distribution, scale & $\epsilon$ & Gaussian, $1\times 10^{-4}$ \\
SZO-$\kappa$ values & $\kappa$ & $\kappa\in\{1,\infty\}$ (sweep in Appendix~\ref{app:kappa-sweep}) \\
Probing mode & & Iterative (simultaneous only in Appendix~\ref{app:seq-vs-sim}) \\
Credit assignment & & Per-sample for NP, NTWP, SZO-1, SZO-$\infty$; per-batch for WP \\
Activation-aware probing &  & Off (Appendix~\ref{app:activation-awareness}) \\
\bottomrule
\end{tabular}

\end{table}

\begin{table}[h]
\centering
\footnotesize
\caption{\textbf{Selected learning rates.} For each method, optimiser and dataset, without and with (bottom block) activation awareness: the learning rate with the lowest mean validation loss over the last $10$ epochs (validation split, $3$ seeds), among grid learning rates at which every seed completed.}
\label{tab:lrs}
\begin{tabular}{lcccccc}
\toprule
 & \multicolumn{3}{c}{Fashion-MNIST} & \multicolumn{3}{c}{CIFAR-10} \\
\cmidrule(lr){2-4}\cmidrule(lr){5-7}
\textbf{Method} & SGD & Adam & SOAP & SGD & Adam & SOAP \\
\midrule
BP & $10^{-2}$ & $3\times10^{-5}$ & $10^{-5}$ & $3\times10^{-2}$ & $3\times10^{-4}$ & $10^{-4}$ \\
NP & $10^{-3}$ & $3\times10^{-5}$ & $10^{-4}$ & $3\times10^{-4}$ & $10^{-4}$ & $10^{-4}$ \\
WP (SZO-0) & $10^{-4}$ & $10^{-4}$ & $10^{-4}$ & $3\times10^{-4}$ & $3\times10^{-4}$ & $3\times10^{-4}$ \\
NTWP & $10^{-3}$ & $3\times10^{-5}$ & $10^{-4}$ & $10^{-3}$ & $10^{-4}$ & $10^{-4}$ \\
SZO-1 & $10^{-3}$ & $3\times10^{-5}$ & $10^{-4}$ & $3\times10^{-3}$ & $3\times10^{-4}$ & $3\times10^{-4}$ \\
SZO-$\infty$ & $10^{-3}$ & $3\times10^{-5}$ & $10^{-4}$ & $3\times10^{-3}$ & $10^{-4}$ & $10^{-4}$ \\
\midrule
\multicolumn{7}{l}{\textit{Activation-aware}} \\
NP & $3\times10^{-2}$ & $3\times10^{-5}$ & $10^{-4}$ & $10^{-3}$ & $3\times10^{-4}$ & $10^{-4}$ \\
WP (SZO-0) & $10^{-4}$ & $10^{-4}$ & $10^{-4}$ & $3\times10^{-3}$ & $3\times10^{-4}$ & $3\times10^{-4}$ \\
NTWP & $10^{-2}$ & $3\times10^{-5}$ & $10^{-4}$ & $3\times10^{-2}$ & $3\times10^{-4}$ & $10^{-4}$ \\
SZO-1 & $3\times10^{-2}$ & $3\times10^{-5}$ & $10^{-4}$ & $10^{-2}$ & $3\times10^{-4}$ & $3\times10^{-4}$ \\
SZO-$\infty$ & $3\times10^{-2}$ & $3\times10^{-5}$ & $10^{-4}$ & $3\times10^{-3}$ & $10^{-4}$ & $3\times10^{-5}$ \\
\bottomrule
\end{tabular}

\end{table}

\newpage

\end{document}